# Unsupervised domain adaptation for cross-modality liver segmentation via joint adversarial learning and self-learning

Jin Hong[1,2], Simon Chun-Ho Yu[1,2], Weitian Chen[1,2,*]

1. CU Lab of AI in Radiology (CLAIR), the Chinese University of Hong Kong, Shatin, Hong Kong SAR, China
2. Department of Imaging and Interventional Radiology, the Chinese University of Hong Kong, Shatin, Hong Kong SAR, China

E-mail: hongj5@mail2.sysu.edu.cn; simonyu@cuhk.edu.hk; wtchen@cuhk.edu.hk;

* Correspondence should be addressed to Weitian Chen

**Abstract:** Liver segmentation on images acquired using computed tomography (CT) and magnetic resonance imaging (MRI) plays an important role in clinical management of liver diseases. Compared to MRI, CT images of liver are more abundant and readily available. However, MRI can provide richer quantitative information of the liver compared to CT. Thus, it is desirable to achieve unsupervised domain adaptation for transferring the learned knowledge from the source domain containing labeled CT images to the target domain containing unlabeled MR images. In this work, we report a novel unsupervised domain adaptation framework for cross-modality liver segmentation via joint adversarial learning and self-learning. We propose joint semantic-aware and shape-entropy-aware adversarial learning with post-situ identification manner to implicitly align the distribution of task-related features extracted from the target domain with those from the source domain. In proposed framework, a network is trained with the above two adversarial losses in an unsupervised manner, and then a mean completer of pseudo-label generation is employed to produce pseudo-labels to train the next network (desired model). Additionally, semantic-aware adversarial learning and two self-learning methods, including pixel-adaptive mask refinement and student-to-partner learning, are proposed to train the desired model. To improve the robustness of the desired model, a low-signal augmentation function is proposed to transform MRI images as the input of the desired model to handle hard samples. Using the public data sets, our experiments demonstrated the proposed unsupervised domain adaptation framework reached four supervised learning methods with a Dice score 0.912 ± 0.037 (mean ± standard deviation).

## 1. Introduction

Deep learning has achieved great success in a variety of medical image analysis tasks in recent years. Ideally, deep neural networks are trained and evaluated on the datasets sampled from the same domain with same underlying distribution. However, this condition is often violated, a phenomenon well-known as domain shift (Kouw and Loog, 2019).

Domain shift often occurs when the images are acquired by different equipment or in different environment (Dou et al., 2018; Gretton et al., 2009; Torralba and Efros, 2011). The domain shift problem also commonly exists in the field of medical image analysis since medical images are often collected from different modalities, scanners, protocols, sites, and population (Novosad et al., 2019; Yang et al., 2019a). Among them, the domain shift due to the difference in modalities is likely the most challenging issue since the imaging physics is different between modalities, resulting in significant mismatch in data distributions (Dou et al., 2018).

To tackle domain shift, domain adaptation technique has been developed to generalize the model trained on the source domain (labeled training dataset) so that it can be applied to the target domain (test dataset) (Patel et al., 2015). Transfer learning can also be used to address this problem. It uses a small amount of training data in the target domain to fine-tune a model that has been well-trained in the source domain so that the model can have accurate prediction in the target domain (Pan and Yang, 2009). However, acquiring high-quality labeled data in target domain is challenging, especially for medical images as they need to be annotated by experienced radiologists. Compared to transfer learning, unsupervised domain adaptation approaches that transfer the knowledge across domains without requiring labeled data in the target domain are desirable. Researchers have investigated unsupervised domain adaptation for segmentation of medical images acquired from different scanners, protocols, and sites. However, the work of cross-modality medical image segmentation is relatively few due to the complexity of the problem (Dou et al., 2018; Yang et al., 2019a).

Accurate liver segmentation of images acquired using computed tomography (CT) or magnetic resonance imaging (MRI) is a prerequisite for clinical applications that require volume and shape measurement of the liver, and monitoring, planning, and virtual or augmented liver surgeries (Howe and Matsuoka, 1999; Van Ginneken et al., 2011). CT and MRI provide complementary diagnostic information of healthy tissues and lesions (Yang et al., 2019a). It is valuable for disease analysis and treatment to provide liver segmentation on both modalities. CT images are more abundant clinically than MR images, but the latter contains richer quantitative diagnostic information (Oliva and Saini, 2004). Thus, it is desirable to achieve unsupervised domain adaptation for transferring the learned knowledge from the source domain containing labeled CT images to the target domain containing unlabeled MR images.

In this paper, we propose a novel unsupervised domain adaptation framework for cross-modality liver segmentation via joint adversarial learning and self-learning. The proposed approach achieved a high performance with a Dice score of $0.912 \pm 0.037$ (mean ± standard deviation). Our method is robust at areas where the contrast between the liver and the background is low. Our main contributions are summarized as follows:

(i) Extended the unsupervised domain adaptation method using joint adversarial learning and self-learning for the semantic segmentation task with unpaired CT and MRI images. The proposed approach achieved comparable performance as four supervised learning-based methods. The code is available at https://github.com/Captain-Hong/UDA-via-joint-adversarial-learning-and-self-learning.

(ii) Proposed a post-situ identification manner for adversarial learning to focus on alignment of task-related features between target and source domains. Such strategy of alignment of partial features reduces the learning difficulties of generator and discriminator models.

(iii) Proposed a joint semantic-aware and shape-entropy-aware adversarial learning to align the distribution of low-level features in latent space extracted from the target and the source domain. By simultaneously using the semantic structure information in logit map as well as the shape and entropy information in weighted self-information map, this approach can ensure learning in a correct direction.

(iv) Proposed a mean completer approach for pseudo-label generation and a low-signal augmentation function to improve the robustness of the segmentation when image contrast or signal is low.

(v) Developed a novel self-learning mechanism named student-to-partner learning and combined it with pixel-adaptive mask refinement for further improving the segmentation results.

## 2. Related work

Unsupervised domain adaptation is a popular research topic in the field of classification and detection, and significant progress has been made in semantic segmentation in recent years. Here we provide a brief review of works of unsupervised domain adaptation on natural images and medical images.

Most research on unsupervised domain adaptation focused on minimizing the distance between the distributions of the features extracted from source and target domains. Tzeng et al. (2014) employed maximum mean discrepancy (MMD) as a distance measure and minimized it with a task-specific loss to achieve domain invariant and semantically meaningful representations. Sun and Saenko (2016) minimized the correlation distance that matches the mean and covariance characteristics of features across target and source domains. These methods developed for classification are limited by the feature representations for distance measure and have difficulties to be used for semantic segmentation due to the complexity of high-dimensional features in these tasks.

With the emergence of generative adversarial network (GAN) (Goodfellow et al., 2014) and its variants (Arjovsky et al., 2017), adversarial loss has been widely used to minimize the domain shift by implicitly learning the mapping between domains. Generally, adversarial learning involves two networks, i.e., a task-specific network for predicting the segmentation map and a discriminator. The task-specific network tries to fool the discriminator by making the features from source and target domains have a similar distribution. Hoffman et al. (2016) firstly introduced adversarial learning approach to unsupervised domain adaptation for semantic segmentation. They applied adversarial learning in a fully convolutional network for aligning global domain by implicitly learning the mapping of feature representations between two domains, and then enabled category specific adaptation by explicit transferring label statistic knowledge from the source to the target domain. Similarly, Chen et al. (2017) used an adversarial learning framework to perform global and class-wise domain alignment with promising results. Ganin et al. (2016) employed two convolutional networks sharing weights to extract domain-invariant features by using adversarial learning. Tzeng et al. (2017) proposed a more flexible framework named adversarial discriminative domain adaptation (ADDA) which combines discriminative modeling, untied weight sharing, and a GAN loss for effective learning in the context of a large domain shift. The aforementioned adversarial learning-based methods mainly use the semantic structural information to align the distributions of features or predictions between source and target domains. Recently, Vu et al. (2019) developed entropy-based unsupervised domain adaptation for semantic segmentation with a competitive performance. They employed an adversarial learning method to minimize the entropy of the prediction map in the target domain to achieve domain adaptation since the model well-trained on the source domain naturally generates the prediction map with low-entropy.

The aforementioned methods implement domain adaptation based on feature-level representations. In contrast, a group of work perform domain adaptation at pixel-level by translating source data to the "style" of the target domain, or vice versa. Bousmalis et al. (2017) proposed a GAN-based method which adapts source-domain images to appear as if they are drawn from the target domain in an unsupervised manner. Zhu et al. (2017) utilized Cycle-GAN to generate the target images according to the images in the source domain for pixel-level adaptation. Hoffman et al. (2018) developed cycle-consistent adversarial domain adaptation (CyCADA) to achieve domain adaptation at both pixel-level and feature-level. Wu et al. (2018) presented dual channel-wise alignment networks (DCAN) to effectively reduce domain shift at both pixel-level and feature-level. Chang et al. (2019) developed a novel domain invariant structure extraction (DISE) framework to disentangle images into domain-invariant structure and domain-specific texture representations to achieve adaptation at pixel-level.

Another group of work focus on self-learning for unsupervised domain adaptation. The idea is to train the current model using the prediction of the previous state of the model or the ensembled model as the pseudo-labels for unlabeled data, which is widely used in semi-supervised approaches (Laine and Aila, 2016; Tarvainen and Valpola, 2017). Zou et al. (2018) developed an iterative self-learning procedure with collaborating class balancing and spatial prior for unsupervised domain adaptation on semantic segmentation. French et al. (2017) proposed self-ensembling for visual domain adaptation problems based on the mean teacher variant (Tarvainen and Valpola, 2017) of temporal ensembling (Laine and Aila, 2016) and achieved state of the art results in many visual benchmarks.

Due to the high cost of medical image annotation and the diversity of data sources, unsupervised domain adaptation is a valuable and important topic in the field of medical image analysis. Kamnitsas et al. (2017) firstly adopted an adversarial learning method to perform unsupervised domain adaptation for brain lesion segmentation in MR images. This study attempted to train a network that is more invariant to differences in input by using a multi-connected domain discriminator. Following this work, Dou et al. (2018) employed adversarial learning to transfer the learned knowledge from the labeled source domain containing MR images to unlabeled target domain containing CT images for cardiac segmentation. They built a domain adaptation module (DAM) to implicitly maps the target input to the latent space of source domain, and used a discriminator named domain critic module (DCM) to identify which domain the features in latent space come from. Chen et al. (2020) developed a novel synergistic image and feature alignment (SIFA) approach to effectively achieve unsupervised domain adaptation between CT and MR images at both pixel-level and feature-level representations. Yang et al. (2019a) achieved unsupervised cross-modality domain adaptation between CT and MR images at pixel-level for liver segmentation by only retaining the domain-invariant content representations via disentangled representation learning. Yang et al. (2019b) proposed a domain-agnostic learning framework with anatomy-consistent embedding (DALACE) to learn a disentangled representation to achieve unsupervised domain adaptation for cross-modality liver segmentation. Perone et al. (2019) extended unsupervised domain adaptation method using self-ensembling based on mean teacher model for spinal cord segmentation in MR images. Novosad et al. (2019) designed an appropriate data augmentation framework to enhance the domain generalization ability of network, and then incorporated a self-ensembling approach to further improve the performance of unsupervised domain adaptation for neuroanatomical segmentation in MR images.

## 3. Method

### 3.1. Overview of the research plan

In order to propose an unsupervised domain adaptation framework for cross-modality liver segmentation, we designed a research plan as shown in Fig. 1. It has 3 major steps, including (i) training source model, (ii) training unsupervised domain adaptation framework, and (iii) testing the desired model. In the first step, an attention U-Net (U1) is trained using labeled CT data in the source domain in a fully-supervised manner. The weights of U1 are shared with another attention U-Net U2. In the second step, we use all labeled CT data and 80% MRI data (unlabeled) to train the proposed unsupervised domain adaptation framework which includes six deep neural networks (U1, U2, U3, U4, D1, and D2). The network U3 with an attention U-Net architecture is the desired model which is used to segment the liver in MR images. In step 3, we validate the performance of the desired model U3 using the rest 20% MR data (labeled). Note we perform the steps 2 and 3 five times to perform 5-fold cross validation.

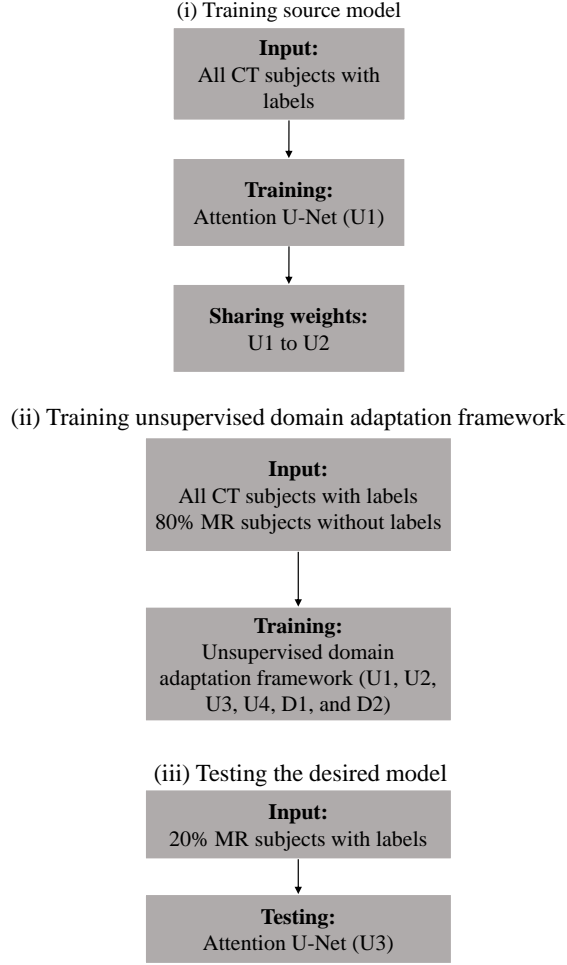

Fig. 1. flowchart of the research plan

## 3.2. The unsupervised domain adaptation framework

In this work, the unsupervised cross-modality domain adaptation task is defined as transferring knowledge from source domain $(X_s, Y_s)$ containing a labeled CT dataset to target domain $X_t$ containing an unlabeled MR dataset. Our proposed framework contains four standard segmentation networks (U1, U2, U3, and U4) and two deep convolutional discriminators (D1 and D2) based on adversarial learning and self-learning, as illustrated in Fig. 2. This framework is designed to provide a well-trained desired model (U3) for segmenting liver from MR images. The other networks are designed to help achieve this goal. When training the framework, the CT images with labels and MR images without labels are simultaneously input into the framework to optimize those segmentation networks and discriminators. Of note, all weights of U1 and part of the weights of U2 were frozen during training, as shown in Fig. 2. In inference, the MR images are input into the desired model U3 for segmentations with only one single forward pass, which is similar to fully-supervised segmentation network.

In our unsupervised domain adaptation framework, six crucial components relevant to our task are proposed: (i) a post-situ identification manner for adversarial learning, (ii) a joint semantic-aware and shape-entropy-aware adversarial learning, (iii) a mean completer of pseudo-label generation, (iv) a low-signal augmentation function, (v) a self-learning method named pixel-adaptive mask refinement, and (vi) a self-learning method named student-to-partner learning. The purpose of the post-situ identification manner for adversarial learning is to avoid aligning the distributions of redundant features. This partial alignment can effectively reduce the learning difficulty of the networks U2, D1 and D2. Based on the post-situ identification manner, we proposed the joint semantic-aware and shape-entropy-aware adversarial learning method to align the distributions of the low-level task-related features extracted from the source domain and the target domain. Since the existence of hard samples (the details of hard samples are described in section 3.6) greatly degrades the performance of the desired model U3, a mean completer of pseudo-label generation and a low-signal augmentation function are designed to improve the robustness of the desired model. The purpose of developing student-to-partner learning and employing pixel-adaptive mask refinement as two self-learning ways is to further improve the segmentation accuracy. The details of these crucial components are provided in the following sections.

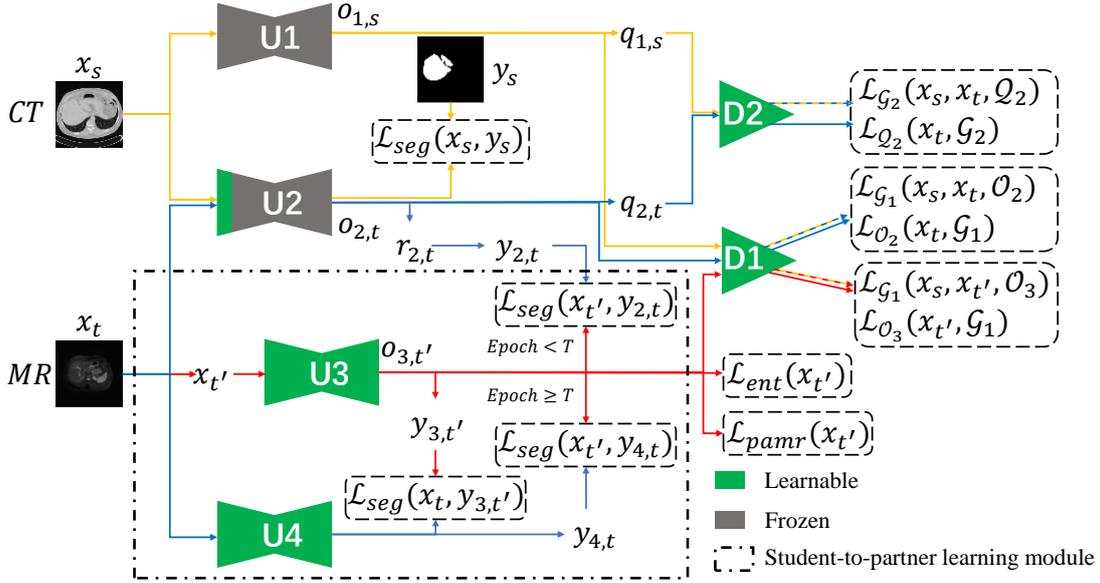

Fig. 2. (Best viewed in color) Overview of the proposed unsupervised domain adaptation framework. Before training the framework, segmentation network U1 is pre-trained on source domain containing a labeled CT dataset, and then the weights of U1 are shared with network U2. The green areas represent the weights of networks that can be updated in training, and the gray areas represent the frozen weights. $x$, $y$, $o$, $r$, and $q$ denote the input image, the label (pseudo-label), logits, recombination of logit maps, and weighted self-information, respectively. $\mathcal{O}$, $\mathcal{Q}$, and $\mathcal{G}$ denote the mapping between the logits and input image, between the weighted self-information and input image, and between the output of discriminator and logits of segmentation network, respectively. $\mathcal{L}$ denotes the loss function.

### 3.3. Segmentation network architecture

In our framework, all segmentation networks (U1, U2, U3, and U4) adopt attention U-Net architecture (Oktay et al., 2018), illustrated in Fig. 3. Note the segmentation networks mainly contain 9 convolution blocks, 4 attention blocks, 4 max-pooling operations, and 4 upsampling operations. Convolution block consisting of 2 stacked convolution layers followed by batch normalization (BN) (Ioffe and Szegedy, 2015). ReLU layers (Nair and Hinton, 2010) is employed to extract representative features. Attention block consisting of gating signal, attention gate and skip connection is used to highlight the features useful for task and suppress the irrelevant regions in input image (Oktay et al., 2018). Max-pooling operation is utilized to downsample the features by factor of 2 at each scale while retaining critical features and achieving invariance to transformations. Upsampling operation consisting of a upsampling layer and a convolution layer that are used to enlarge the feature maps by a factor of 2 and smooth out them. In the encoding part, the number of channels is increased by a factor of 2 every time the convolution block is performed. On the contrary, in the decoding part, the number of channels is reduced by a factor of 2 every time the upsampling operation is performed.

Considering a labeled dataset of $N$ samples, denoted as $(X,Y) = \{(x^1,y^1),\cdots,(x^N,y^N)\}$, segmentation network is used to learn a mapping from the input images $X$ to the labels $Y$. In our task, $x^i$ represents the medical image and $y^i$ represents the anatomical structure. The input image $x^i$ alternately forwards to 5 convolution blocks and 4 downsampling operations in the encoding part of the network for extracting task-related feature maps. Afterwards, the feature maps alternately pass through 4 upsampling operations, 4 attention blocks and 4 convolution blocks in the decoding part of the network for outputting logit map $o^i$ with the same size as the input. Finally, sigmoid function is used to calculate the probability of each pixel in the logit map for prediction map $p^i$.

The segmentation network is optimized by minimizing the segmentation loss $\mathcal{L}_{seg}$ between the label $y^i$ and the prediction map $p^i$. In our study, dice coefficient is employed as the segmentation loss to avoid the situations that the predictions are strongly biased towards background since the anatomy of interest occupies only small regions in the images (Milletari et al., 2016). Note the segmentation loss can be expressed as follows:

$$\mathcal{L}_{seg}(X,Y) = \frac{\sum_{i=1}^{N} 2y^i p^i}{\sum_{i=1}^{N} y^i y^i + \sum_{i=1}^{N} p^i p^i}, \qquad (1)$$

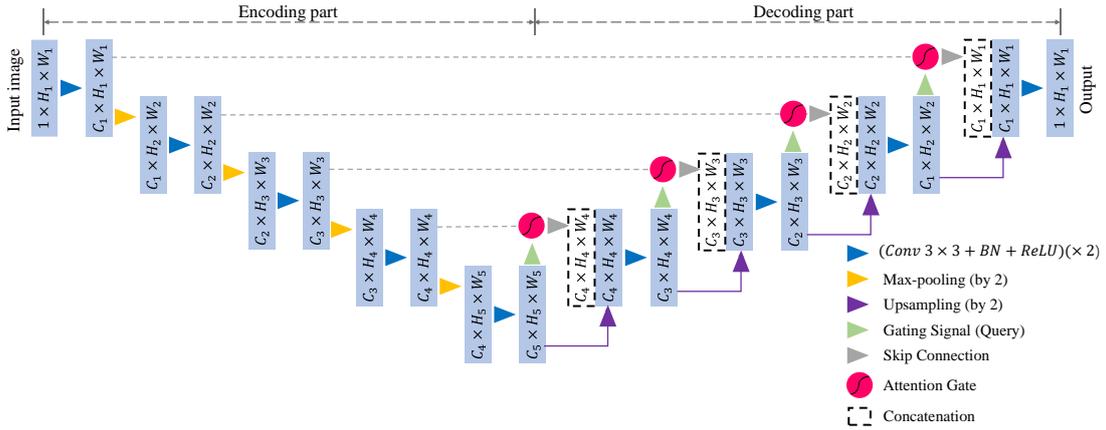

Fig. 3. (Best viewed in color) The proposed segmentation networks architecture (attention U-Net(Oktay et al., 2018)). $C$, $H$, and $W$ represent the number of channels, the height, and the width of feature maps respectively.

### 3.4. Post-situ identification for adversarial learning

To achieve unsupervised domain adaptation, we define two types of adversarial learning for aligning the distributions of features extracted from the source domain and the target domain according to the relative position between the features and the discriminator. The first adversarial learning has an in-situ identification manner when the discriminator immediately follows the features. The second adversarial learning has a post-situ identification manner when the discriminator is far behind the features. The design is illustrated in Fig. 4.

Note the segmentation network U1 is pre-trained on the labeled CT dataset by fully-supervised learning with its weights shared with U2. Only a part of weights of U2 will be updated in training. This is indicated by green area in Fig. 4. The distribution of features $f_{2,t}$ extracted from the target domain are first aligned with the distribution of features $f_{1,s}$ extracted from the source domain, and then input into the rest part of U2 to achieve accurate segmentation of MR images. Note U1, U2 and D3 constitute the in-situ identification manner for adversarial learning, while U1, U2 and D1 constitute the post-situ identification manner.

To determine how well the two distributions of $f_{1,s}$ and $f_{2,t}$ are aligned, discriminator D1/D3 is employed to implicitly estimate the distance between them. In the in-situ identification manner, $f_{1,s}$ and $f_{2,t}$ are directly input into discriminator D3 for measuring the distance. In post-situ identification manner, those features are input into the filters (here we regard the rest part of the network as a filter) firstly for logits ($o_{1,s}$ and $o_{2,t}$) which are then input into discriminator D1. Thus, the distance between the two distributions of $f_{1,s}$ and $f_{2,t}$ is indirectly measured from the corresponding distributions of logits $o_{1,s}$ and $o_{2,t}$.

The task of D3/D1 is to accurately identify which domain the features ($f_{1,s}$ or $f_{2,t}$) or logits ($o_{1,s}$ or $o_{2,t}$) belong to, while the task of U2 is to fool the discriminator by making the distribution of $f_{2,t}$ close to that of $f_{1,s}$. Therefore, the combination of segmentation network U1 and U2, which is considered as the generator model, and the discriminator model D1/D3 form a minimax two-player game, same as GAN (Goodfellow et al., 2014).

The mapping between the source domain input $x_s$ and features $f_{1,s}$ (or logits $o_{1,s}$) is termed as $\mathcal{F}_1$ (or $\mathcal{O}_1$), and the mapping between the target domain input $x_t$ and features $f_{2,t}$ (or logits $o_{2,t}$) is termed as $\mathcal{F}_2$ (or $\mathcal{O}_2$). Given the distribution of the source domain features $f_{1,s} = \mathcal{F}_1(x_s) \sim \mathbb{P}_{f,s}$, and that of the target domain features $f_{2,t} = \mathcal{F}_2(x_t) \sim \mathbb{P}_{f,t}$, the distance between the two distributions that requires to be minimized is denoted $W(\mathbb{P}_{o,s}, \mathbb{P}_{o,t})$. Here the Wasserstein distance (Earth-Mover distance) (Arjovsky et al., 2017) between the two distributions is adopted for stabilized training. Its definition is given as follows:

$$W(\mathbb{P}_{f,s}, \mathbb{P}_{f,t}) = \inf_{\gamma \sim \prod(\mathbb{P}_{f,s}, \mathbb{P}_{f,t})} \mathbb{E}_{(x,y) \sim \gamma}[\|x - y\|], \tag{2}$$

where $\prod(\mathbb{P}_{f,s}, \mathbb{P}_{f,t})$ denotes the set of all joint distributions $\gamma(x, y)$ whose marginals are respectively $\mathbb{P}_{f,s}$ and $\mathbb{P}_{f,t}$.

Similarly, the Wasserstein distance between the two distributions of logits is given as follows:

$$W(\mathbb{P}_{o,s}, \mathbb{P}_{o,t}) = \inf_{\gamma \sim \prod(\mathbb{P}_{o,s}, \mathbb{P}_{o,t})} \mathbb{E}_{(x,y) \sim \gamma}[\|x - y\|], \tag{3}$$

The mapping between the features ($f_{1,s}$ or $f_{2,t}$) and the output of D3 is denoted as $\mathcal{G}_3$, and the mapping between the logits ($o_{1,s}$ or $o_{2,t}$) and the output of D1 is denoted as $\mathcal{G}_1$. In adversarial learning, we jointly optimize the generator U2 and the discriminator D1/D3 via adversarial losses. Given the target dataset $X_t$, the loss for optimizing the generator model U2 in in-situ identification manner can be represented as follows:

$$\min_{\mathcal{F}_2} \mathcal{L}_{\mathcal{F}_2}(X_t, \mathcal{G}_3) = -\mathbb{E}_{\mathcal{F}_2(x_t) \sim \mathbb{P}_{f,t}}[\mathcal{G}_3(\mathcal{F}_2(x_t))], \tag{4}$$

Given the source dataset $X_s$, the discriminator model D3 is optimized under the loss:

$$\min_{\mathcal{G}_3} \mathcal{L}_{\mathcal{G}_3}(X_s, X_t, \mathcal{F}_2) = \mathbb{E}_{\mathcal{F}_2(x_t) \sim \mathbb{P}_{f,t}}[\mathcal{G}_3(\mathcal{F}_2(x_t))] - \mathbb{E}_{\mathcal{F}_1(x_s) \sim \mathbb{P}_{f,s}}[\mathcal{G}_3(\mathcal{F}_1(x_s))], \; s.t. \|\mathcal{G}_3\|_{L \leq K}, \tag{5}$$

where $K$ is a constant for applying Lipschitz constraint to $\mathcal{G}_3$.

Similarly, the losses for optimizing the generator model U2 and the discriminator D1 in post-situ identification manner can be given respectively as follows:

$$\min_{\mathcal{O}_2} \mathcal{L}_{\mathcal{O}_2}(X_t, \mathcal{G}_1) = -\mathbb{E}_{\mathcal{O}_2(x_t) \sim \mathbb{P}_{o,t}}[\mathcal{G}_1(\mathcal{O}_2(x_t))], \tag{6}$$

$$\min_{\mathcal{G}_1} \mathcal{L}_{\mathcal{G}_1}(X_s, X_t, \mathcal{O}_2) = \mathbb{E}_{\mathcal{O}_2(x_t) \sim \mathbb{P}_{o,t}}[\mathcal{G}_1(\mathcal{O}_2(x_t))] - \mathbb{E}_{\mathcal{O}_1(x_s) \sim \mathbb{P}_{o,s}}[\mathcal{G}_1(\mathcal{O}_1(x_s))], \ s.t. \|\mathcal{G}_1\|_L \leq K, \quad (7)$$

The weights of the generator model U2 and the discriminator model D1/D3 are alternately updated during training. With the updating of the weights, the Wasserstein distance is estimated more and more accurately by discriminator, and the generator is more effective to produce source-like features/logits for achieving unsupervised cross-modality domain adaptation.

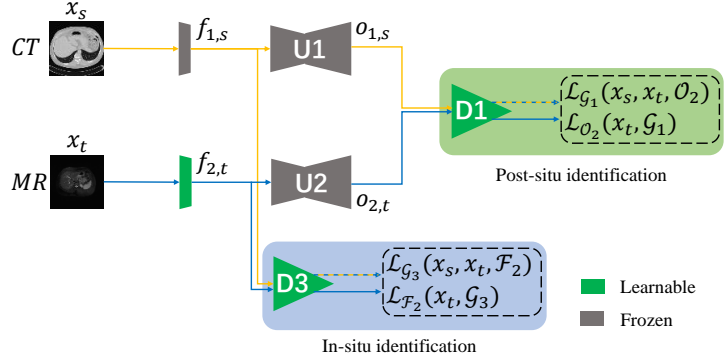

Fig. 4. (Best viewed in color) In-situ and post-situ identification manners for adversarial learning for unsupervised domain adaptation.

### 3.5. Semantic-aware and shape-entropy-aware adversarial learning

Note the part consisting of U1, U2 and D1 in the Fig. 2 is considered as semantic-aware adversarial learning for unsupervised domain adaptation since D1 uses the logit map containing rich semantic structural information as input. Fig. 5 (c) gives eight logit maps output by the segmentation network U1 pre-trained on source CT dataset in a fully-supervised learning manner. Note logit maps contain not only clear structural information of the liver, but also structural information of other organs. In addition to the overall semantic structural information given by logit maps, the shape and entropy information of prediction maps also carry important information to train the model (Chung et al., 2021; He et al., 2021; Vu et al., 2019).

Interestingly, it is noticed that weighted self-information map (Vu et al., 2019) can characterize the shape information of the liver while calculating the entropy information of prediction map in our task. As Fig. 5 (d) shows, the bright point cloud in the weighted self-information map clearly outlines the shape of the liver. Each pixel represents the confidence of prediction with dark and bright pixels corresponding to high confidence (low-entropy) and low confidence (high-entropy), respectively.

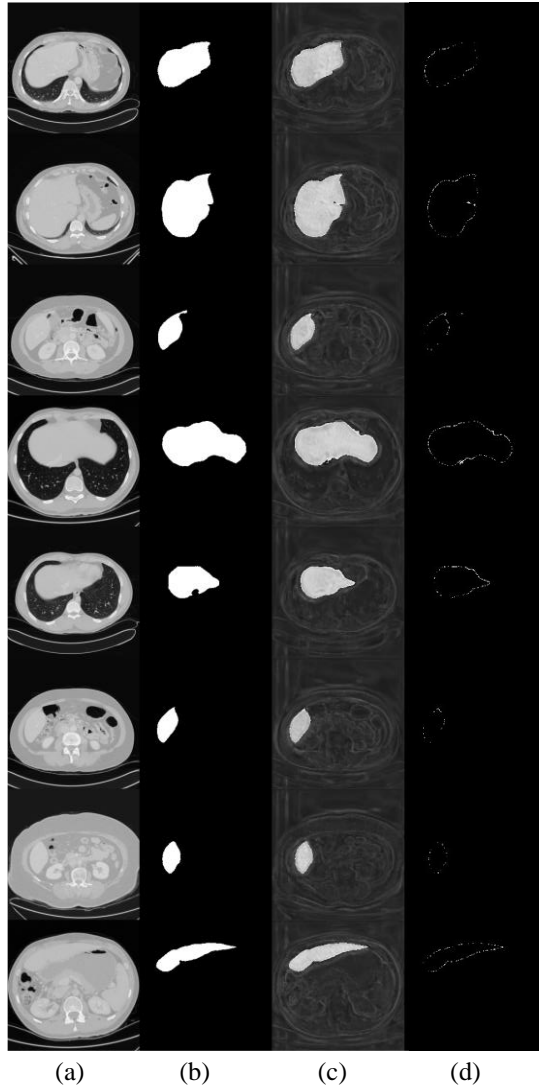

(a) (b) (c) (d)

Fig. 5. (Best viewed in zooming in) Examples of logit map and weighted self-information map output from the segmentation network U1 pre-trained on source CT dataset under fully-supervised learning. (a) CT image; (b) Ground truth; (c) logit map; (d) weighted self-information map.

To illustrate the origin of the weighted self-information map, we take the output of segmentation network U1 in the Fig. 2 as an example. Given a pixel $o_{1,s}^{h,w}$ from the $H \times W$ logit map $o_{1,s}$, the prediction is expressed as $p_{1,s}^{h,w} = sigmoid(o_{1,s}^{h,w})$. The self-information can be represented as $-\log p_{1,s}^{h,w}$ (Katona and Nemetz, 1976) and the weighted self-information is defined as $q_{1,s}^{h,w} = -p_{1,s}^{h,w} \cdot \log p_{1,s}^{h,w}$. All the pixels with weighted self-information together make up the map $q_{1,s}$. In order to further understand the meaning of weighted self-information map, we plot the curve of weighted self-information versus probability in Fig. 6. Note that when the probability is close to 0 or 1, the weighted self-information value will be close to 0, shown as black pixels in Fig. 5 (d). On the contrary, when the probability is away from 0 or 1, the weighted self-information value will be greater than 0, shown as white pixels in Fig. 5 (d). In general, the prediction probability of the boundaries between the liver and the background is far away from 0 or 1 and close to 0.5 (0.5 means that the prediction result has the lowest confidence), while the prediction probability of other places is close to 0 or 1 (0 and 1 mean that the prediction result has the highest confidence). Thus, the weighted self-information map can highlight the boundaries (shape information) of the liver, as shown in Fig. 5 (d).

The shape and entropy information are simultaneously used in the part of framework consisting of U1, U2 and D2 in the Fig. 2. Thus, we name it as shape-entropy-aware adversarial learning for unsupervised domain adaptation. The mapping between source domain input $x_s$ and weighted self-information map $q_{1,s}$ is termed $Q_1$, and the mapping between target domain input $x_t$ and weighted self-information map $q_{2,t}$ is termed $Q_2$. The mapping between weighted self-information map ($q_{1,s}$ or $q_{2,t}$) and the output of D2 is denoted by $\mathcal{G}_2$. The losses for optimizing the generator model U2 and the discriminator model D2 can be given respectively as follows according to formulas (6) and (7):

$$\min_{Q_2} \mathcal{L}_{Q_2}(X_t, \mathcal{G}_2) = -\mathbb{E}_{Q_2(x_t) \sim \mathbb{P}_{q,t}}[\mathcal{G}_2(Q_2(x_t))], \tag{8}$$

$$\min_{\mathcal{G}_2} \mathcal{L}_{\mathcal{G}_2}(X_s, X_t, Q_2) = \mathbb{E}_{Q_2(x_t) \sim \mathbb{P}_{q,t}}[\mathcal{G}_2(Q_2(x_t))] - \mathbb{E}_{Q_1(x_s) \sim \mathbb{P}_{q,s}}[\mathcal{G}_2(Q_1(x_s))], \ s.t. \|\mathcal{G}_2\|_{L \leq K}, \tag{9}$$

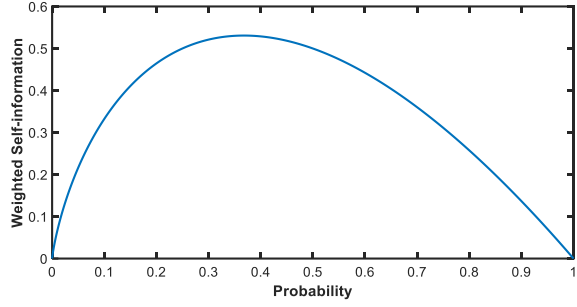

Fig. 6. Curve of weighted self-information versus probability

### 3.6. Mean completer of pseudo-label generation

As Fig. 2 shows, we generate three kinds of pseudo-label in the proposed unsupervised domain adaptation framework denoted as $y_{2,t}$, $y_{3,t'}$ and $y_{4,t}$, respectively. The $y_{2,t}$, $y_{3,t'}$ and $y_{4,t}$ are generated based on the outputs of the segmentation networks U2, U3 and U4, respectively, with MR images as the input. Among them, $y_{3,t'}$ and $y_{4,t}$ are generated via a normal manner. The logit map $o_{3,t'}$ ($o_{4,t}$) is input into sigmoid function to generate prediction map $p_{3,t'}$ ($p_{4,t}$). The threshold of 0.5 is used to group those predictions into 0 or 1 (0 denotes background and 1 denotes liver).

Under certain conditions, liver MRI has very low signal and is close to the background. This happens when the patient has liver iron overload as the iron shortens the transverse relaxation time of the liver significantly which results in rapid signal decay in MRI. Consequently, the pseudo-label $y_{2,t}$ cannot be properly generated via the normal manner since U2 cannot detect any pixel belonging to liver during training, as shown in Fig. 7.

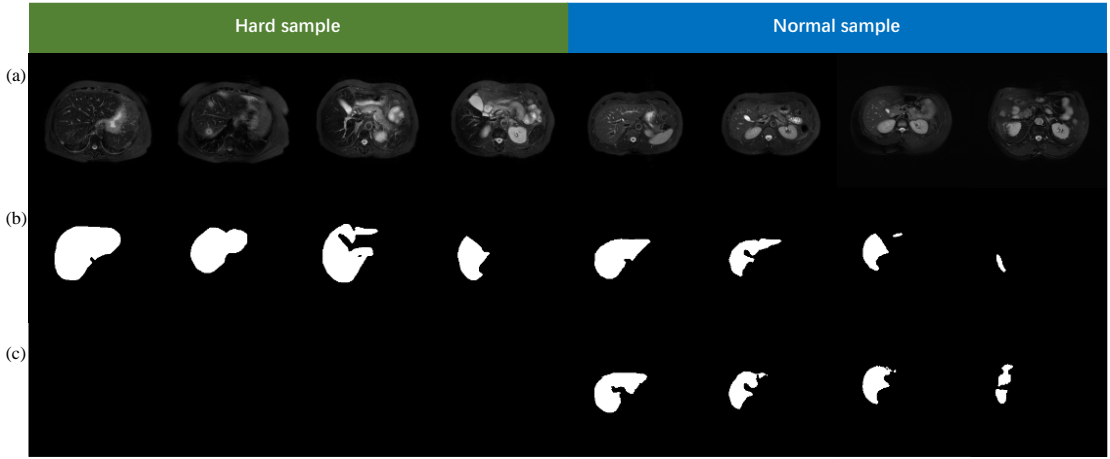

Fig. 7. Examples of pseudo-label generated via the normal manner based on the logit map outputted by the segmentation network U2 during training. The green area is the hard sample, and the blue area is the normal sample. (a) MR image; (b) Ground truth; (c) Pseudo-label.

Inspired by mean completer, one of the traditional data completion methods (Tong et al., 2020), we proposed the mean completer of pseudo-label generation to address this problem. Given a mini-batch with $n$ samples, we generate all pseudo-labels of the samples by the normal manner based on the logit maps $\{o_{2,t}^1, o_{2,t}^2, \cdots, o_{2,t}^n\}$. Here the sample is considered as hard sample when all pixels in its corresponding pseudo-label are 0. We then calculate the average logit map $o_{2,t}^{ave} = (o_{2,t}^1 + o_{2,t}^2 + \cdots + o_{2,t}^n)/n$ in an element-wise manner. After that, the logit maps of hard samples are replaced by the average logit map $o_{2,t}^{ave}$ and that of normal samples are retained. The new recombination of logit maps $\{r_{2,t}^1, r_{2,t}^2, \cdots, r_{2,t}^n\}$ are obtained and the new pseudo-labels $\{y_{2,t}^1, y_{2,t}^2, \cdots, y_{2,t}^n\}$ can now be obtained via the normal manner.

Compared with the normal manner, the mean completer of pseudo-label generation can provide a more useful cue for U3 when a mini-batch of input contains hard samples. When the average logit map is converted to pseudo-label via normal manner, three situations can appear theoretically, showing in Fig. 8 (c1), (c2) and (c3), respectively. The first situation (c1) means that one or more foreground parts of logit maps are retained, while the second situation (c2) means that the overlapped regions are retained. The third situation (c3) means none is retained, which rarely happens. As Fig. 8 (d1) or (d2) shows, the generated pseudo-label in the first or second situation generally contains useful information (pixels which are overlapped with the ground truth) more or less for supervising U3.

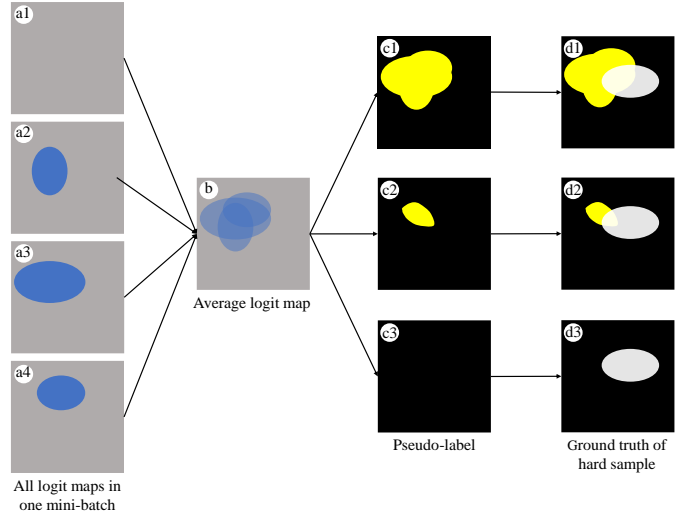

Fig. 8. (Best viewed in color) A toy example of mean completer of pseudo-label generation. (a1) denotes the logit map of hard sample. (a2), (a3) and (a4) denote the logit maps of normal samples. (c1), (c2) and (c3) represent three possible pseudo-labels. (d1), (d2) and (d3) represent the possible spatial relationships between ground truth and the above three pseudo-labels.

### 3.7. Low-signal augmentation function

To further improve the robustness of U3, we propose a low-signal augmentation function to transform all samples to stretch the distance between the pixel value of the liver and that of low-signal background. Such transformation process also makes the hard samples closer to the normal samples. Given an MR image $x_t$ with size of $H \times W$, the transformed image $x_{t'}$ considered as the input of U3 can be calculated in an element-wise manner according to the proposed low-signal augmentation function as follows:

$$z_t = \log(x_t) + \beta x_t \tag{10}$$

$$x_{t'} = \left( \frac{z_t - \min_{(i,j) \in H \times W} z_t^{i,j}}{\max_{(i,j) \in H \times W} z_t^{i,j} - \min_{(i,j) \in H \times W} z_t^{i,j}} \right)^2 \tag{11}$$

where $\beta$ is a trade-off parameter and is a natural number.

Fig. 9 gives the curve of low-signal augmentation function with different trade-off parameters $\beta$. Note the transformed value increases sharply in the low value region and increase slowly in the high value region. With the decrease of $\beta$, the distance of pixel values in the low value region increases, while that in the high value region shows an opposite trend. Fig. 10 shows results after applying this function to MR images. Note this function significantly increases the contrast between regions with low signals in the liver and the background. With the decrease of $\beta$, the noise level increases and the contrast between the liver and other organs decreases. Therefore, the parameter $\beta$ needs to be determined by experiments according to the specific situation.

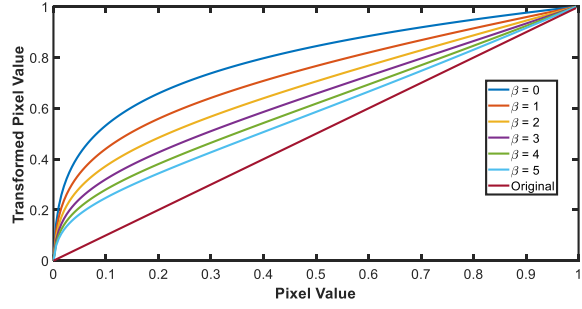

Fig. 9. (Best viewed in color) Curve of the proposed low-signal augmentation function with different trade-off parameters $\beta$.

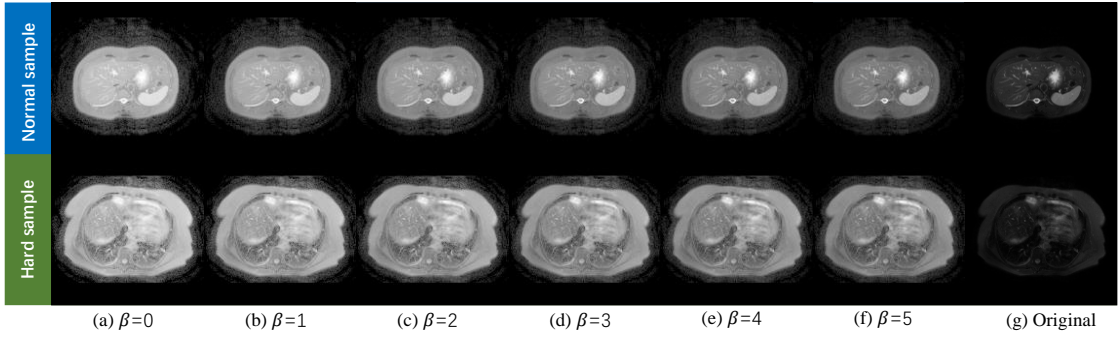

(a) $\beta=0$  (b) $\beta=1$  (c) $\beta=2$  (d) $\beta=3$  (e) $\beta=4$  (f) $\beta=5$  (g) Original

Fig. 10. Effect of low-signal augmentation with different trade-off parameters $\beta$ on MR image.

### 3.8. Pixel-adaptive mask refinement

Local consistency is defined to measure whether neighboring pixels with similar appearance are assigned to the same class in Araslanov and Roth (2020). Pixel-adaptive mask refinement was proposed to achieve high local consistency for semantic mask (prediction map). Similar to other mask refinement methods, such as GrabCut (Rother et al., 2004) and dense CRFs (Krähenbühl and Koltun, 2012), the proposed pixel-adaptive mask refinement can provide self-supervision for training the segmentation model and achieving more accurate predictions (Araslanov and Roth, 2020). This method overcomes the side effect of long training time when conventional mask refinement methods are adopted.

Fig. 11 illustrates pixel-adaptative mask refinement. Given the pixel-level semantic mask $p_{:,:,:} \in (0,1)^{C \times H \times W}$ ("$C$" denotes $C$ classes including $(C-1)$ object classes and 1 background class) and the image $x$, the main idea of pixel-adaptive mask refinement derived from pixel-adaptive convolution (Su et al., 2019) is to update each pixel label $p_{:,i,j}$ iteratively with a convex combination of its neighbors $\mathcal{N}(i,j)$. For example, the pixel label $p_{:,i,j}$ at $m^{\text{th}}$ iteration will be refined as follows:

$$p_{:,i,j}^m = \sum_{(l,n) \in \mathcal{N}(i,j)} \alpha_{i,j,l,n} \cdot p_{:,l,n}^{m-1}, \qquad (12)$$

where the pixel-level affinity $\alpha_{i,j,l,n}$ is a function about the image $x$. A kernel function $k$ on the pixel intensities $x$ is employed to compute $\alpha_{i,j,l,n}$,

$$k(x_{i,j}, x_{l,n}) = -\frac{x_{i,j} - x_{l,n}}{\sigma_{i,j}^2}, \qquad (13)$$

where $\sigma_{i,j}$ denotes the standard deviation of the image intensity computed locally for the affinity kernel. Afterwards, $\alpha_{i,j,l,n}$ can be obtained as follows:

$$\alpha_{i,j,l,n} = \frac{e^{\bar{k}(x_{i,j}, x_{l,n})}}{\sum_{(q,r) \in \mathcal{N}(i,j)} e^{\bar{k}(x_{i,j}, x_{q,r})}}, \qquad (14)$$

where $\bar{k}$ denotes the average affinity value over the image channels.

Pixel-adaptative mask refinement is considered as a parameter-free recurrent module for local refinement, updating the labels iteratively following formula (12). It is obvious that the number of required iterations depends on the affinity kernel size and shape (e.g., 3×3 in Fig. 11). Multiple kernels with different dilation rates can achieve a better performance in practice. More details about pixel-adaptive mask refinement can be found in Araslanov and Roth (2020).

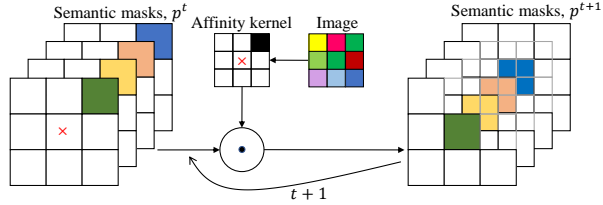

Fig. 11. (Best viewed in color) Illustration for pixel-adaptive mask refinement(Araslanov and Roth, 2020). The proximity of each pixel to its neighbors, which is denoted as affinity kernel, in channel space of input image is computed firstly. Afterwards, the kernel is applied to the semantic masks iteratively with using an adaptive convolution for the refined masks.

In this paper, we borrowed this method as a self-learning method to further improve the segmentation performance of the desired model U3. With the semantic mask $p_{3,t'} = sigmoid(o_{3,t'})$ and the input image $x_{t'}$ of U3 (Fig. 2), the refined semantic mask $p_{3,t',re}$ can be obtained by pixel-adaptive mask refinement method and its pseudo-label $y_{3,t',re}$ can also be obtained. Given an unlabeled dataset containing $N$ MR images, the refinement (self-learning) loss $\mathcal{L}_{pamr}$ provided by pixel-adaptive mask refinement for supervising U3 is given as below:

$$\mathcal{L}_{pamr}(X_{t'}) = \mathcal{L}_{seg}(X_{t'}, Y_{3,t',re}) = \frac{\sum_{i=1}^{N} 2 y_{3,t',re}^i p_{3,t'}^i}{\sum_{i=1}^{N} y_{3,t',re}^i y_{3,t',re}^i + \sum_{i=1}^{N} p_{3,t'}^i p_{3,t'}^i}, \tag{15}$$

**3.9. Student-to-partner learning**

To further improve the performance of the desired model U3, we extend the advantages of traditional knowledge distillation (Hinton et al., 2015) and deep mutual learning (Zhang et al., 2018) to unsupervised domain adaptation and propose a two-stage learning mechanism named student-to-partner learning. Fig. 12(a) and 11(b) illustrates the core idea of deep mutual learning (Zhang et al., 2018) and the proposed student-to-partner learning, respectively. Of note, the Kullback Leibler (KL) divergence of the distributions of predictions is used as the mimicry loss in Zhang et al. (2018). Here the segmentation loss (Dice loss) ($\mathcal{L}_{seg}(x, y_1) + \mathcal{L}_{seg}(x, y_2)$) provided by pseudo-labels $y_1$ and $y_2$ is employed as the mimicry loss to avoid the situation that the predictions are strongly biased towards the background. In student-to-partner learning, the network S1 is used as the teacher model and the network S2 as the student model at the first stage ($Epoch < T$). S1 learns from the input image $x$ under the supervision of pseudo-label $y$, while S2 learns from the transformed image $x'$ under the supervision of the pseudo-label $y_1$ produced by S1. After these two models are fully trained, the segmentation loss provided by the pseudo-label $y$ will be abandoned. These two models then enter the second stage ($Epoch \geq T$) and become partners to learn from each other. In our design of student-to-partner learning, the two models have different learning process, different input, and different network structure. Those differences coupled with the randomness of the deep neural network learning itself ensure the two models learn different knowledge, which is conducive to the mutual learning during the second stage. The overall loss $\mathcal{L}_{stpl}$ of student-to-partner learning is given below:

$$\mathcal{L}_{stpl} = \begin{cases} \mathcal{L}_{seg}(X',Y_1) + \mathcal{L}_{seg}(X,Y) & if\ Epoch < T \\ \mathcal{L}_{seg}(X',Y_1) + \mathcal{L}_{seg}(X,Y_2) & if\ Epoch \geq T \end{cases}, \qquad (16)$$

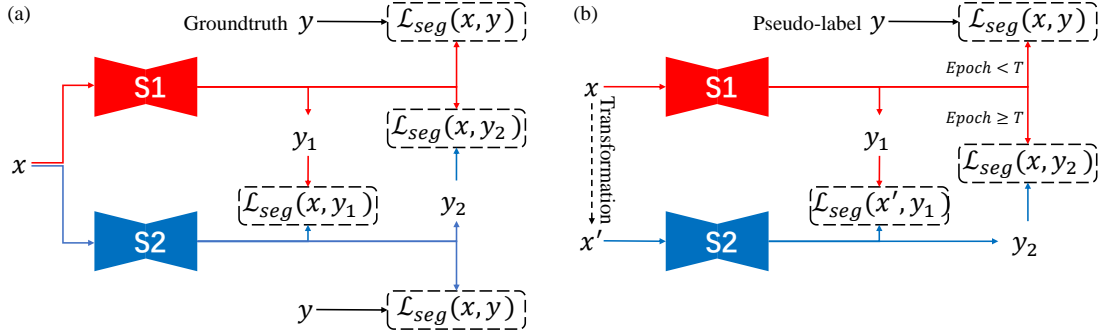

Fig. 12. (Best viewed in color) The illustration of deep mutual learning and student-to-partner learning. (a) denotes the deep mutual learning with the pseudo-label-based mimicry loss, (b) denotes the proposed student-to-partner learning.

We employ the student-to-partner learning as a self-learning method to improve the performance of the desired model U3, as shown in Fig. 2. The segmentation network U3 and U4 constitute the student-to-partner learning mechanism. The pseudo-label $y_{3,t'}$ produced by the teacher model U3 supervised by the pseudo-label $y_{2,t}$ is used for supervised learning of the student model U4 to segment liver from MR image at the first stage. After training the teacher and student models with epoch greater than the set value, the pseudo-label $y_{2,t}$ is abandoned and the two models start to learn from each other, called the second stage.

### 3.10. Overall loss function and technical details

To train the proposed unsupervised domain adaptation framework (see Fig. 2), all losses are combined together for optimization. The overall loss $\mathcal{L}_{all}$ is as follows:

$$\mathcal{L}_{all} = \begin{cases} \begin{aligned} &\lambda_1 \mathcal{L}_{seg}(X_s,Y_s) + \lambda_2 \mathcal{L}_{seg}(X_{t'},Y_{2,t}) + \\ &\lambda_3 \mathcal{L}_{seg}(X_t,Y_{3,t'}) + \lambda_4 \mathcal{L}_{pamr}(X_{t'}) + \\ &\lambda_5 \mathcal{L}_{\mathcal{D}_1}(X_s,X_t,\mathcal{O}_2) + \lambda_6 \mathcal{L}_{\mathcal{O}_2}(X_t,\mathcal{D}_1) + \\ &\lambda_7 \mathcal{L}_{\mathcal{D}_1}(X_s,X_{t'},\mathcal{O}_3) + \lambda_8 \mathcal{L}_{\mathcal{O}_3}(X_{t'},\mathcal{D}_1) + \\ &\lambda_9 \mathcal{L}_{\mathcal{D}_2}(X_s,X_t,\mathcal{Q}_2) + \lambda_{10} \mathcal{L}_{\mathcal{Q}_2}(X_t,\mathcal{D}_2) + \lambda_{11} \mathcal{L}_{ent}(X_{t'}) \end{aligned} & if\ Epoch<T \\ \\ \begin{aligned} &\lambda_1 \mathcal{L}_{seg}(X_s,Y_s) + \lambda_2 \mathcal{L}_{seg}(X_{t'},Y_{4,t}) + \\ &\lambda_3 \mathcal{L}_{seg}(X_t,Y_{3,t'}) + \lambda_4 \mathcal{L}_{pamr}(X_{t'}) + \\ &\lambda_5 \mathcal{L}_{\mathcal{D}_1}(X_s,X_t,\mathcal{O}_2) + \lambda_6 \mathcal{L}_{\mathcal{O}_2}(X_t,\mathcal{D}_1) + \\ &\lambda_7 \mathcal{L}_{\mathcal{D}_1}(X_s,X_{t'},\mathcal{O}_3) + \lambda_8 \mathcal{L}_{\mathcal{O}_3}(X_{t'},\mathcal{D}_1) + \\ &\lambda_9 \mathcal{L}_{\mathcal{D}_2}(X_s,X_t,\mathcal{Q}_2) + \lambda_{10} \mathcal{L}_{\mathcal{Q}_2}(X_t,\mathcal{D}_2) + \lambda_{11} \mathcal{L}_{ent}(X_{t'}) \end{aligned} & if\ Epoch \geq T \end{cases}, \qquad (17)$$

where all $\lambda$ denote the weights that are used to balance those losses. When epoch is less than $T$, $\lambda_1$, $\lambda_2$, $\lambda_3$, $\lambda_4$, $\lambda_6$, $\lambda_8$, and $\lambda_{10}$ are set as 1, $\lambda_5$, $\lambda_7$, and $\lambda_9$ are set as 0.5, and $\lambda_{11}$ is set as 5. When epoch is bigger than or equal to $T$, the parameter $\lambda_2$ is changed to 5 and the others remain unchanged. $T$ should be set large enough to train all networks (U2, U3, U4, D1 and D2) fully in the whole framework. $T$ was set as 70 in our task.

All the mathematical formulas of losses in our framework are mentioned above, except for the entropy minimization loss $\mathcal{L}_{ent}(X_{t'})$ which is used to suppress the high entropy probability predictions of the desired model U3. Given an unlabeled dataset $X_{t'}$ of $N$ transformed MR images and a pixel $o_{3,t'}^{h,w}$ from the $H \times W$ logit map $o_{3,t'}$, the probability prediction is expressed as $p_{3,t'}^{h,w} =$

$sigmoid(o_{3,t'}^{h,w})$. The entropy minimization loss $\mathcal{L}_{ent}(X_{t'})$ is defined as follows:

$$\mathcal{L}_{ent}(X_{t'}) = \frac{\sum_{n=1}^{N}\sum_{h=1}^{H}\sum_{w=1}^{W}(-p_{3,t'}^{n,h,w}\log(p_{3,t'}^{n,h,w}))}{N}, \quad (18)$$

Note all our segmentation networks adopt attention U-Net architecture (Fig. 3). The only difference between these networks is the number of filters in their structure. The numbers of filters in the first convolution block of U1, U2, U3, and U4 are set as 64, 64, 32, and 8, respectively. The numbers of filters in the following convolution blocks can be deduced according to the previous introduction (see section 3.3). The outputs of the first convolution block in U1 and U2, which is denoted by the first $C_1 \times H_1 \times W_1$ in Fig. 3, are set as the features whose distributions need to be aligned. We construct the discriminator networks D1 and D2 according to the discriminator structure in deep convolutional GAN (DCGAN) (Radford et al., 2015). D1 and D2 have exactly the same structure, containing 7 convolution layers, each (except for the last convolution layer) followed by a batch normalization layer and a leaky-ReLU (Liew et al., 2016) layer with a fixed negative slope of 0.2. The trade-off parameter $\beta$ in low-signal augmentation function is set as 3 which is determined by the grid search method. In terms of pixel-adaptative mask refinement, the iterations, the affinity kernel size, and composition of dilation rates are set as 10, 3×3, [1,2,4,8], respectively, determined by grid search method.

## 4. Experiments and Results

### 4.1. Dataset

The experiments in this study were performed on two public challenge datasets of the LiTS-Liver Tumor Segmentation Challenge released with ISBI 2017 (Bilic et al., 2019) and the CHAOS-Combined (CT-MR) Healthy Abdominal Organ Segmentation released with ISBI 2019 (Kavur et al., 2019). LiTS dataset consists of 131 subjects with CT images and CHAOS dataset consists of unpaired 20 subjects with CT images and 20 subjects with MR images (T1-DUAL and T2-SPIR). All patients in LiTS have liver cancer (pixels belonging to tumors in liver are considered as liver in our experiments), while all patients in CHAOS are potential liver donors having healthy liver (no tumors, lesions or any other diseases). All CT images were resized to the same size as MR images with matrix size of 256×256. The pixel value of CT images was limited within -1000 to 400 (the overflow value is taken as the boundary value), while that of MR images was not limited. Both CT and MR images were normalized to 0 to 1. 151 (131+20) subjects with CT images and 20 subjects with MR images (T2-SPIR) are set as the source domain and the target domain respectively in this work. The MR images are randomly divided into 5 folds (each fold contains 4 subjects) for performing subject-wise cross-validation. Of note, we abandoned the background slices from each subject and only retained the liver-contained images in our study, which results in total 21552 slices in CT domain and 408 slices in MR domain.

### 4.2. Evaluation metrics

Seven commonly used metrics in the field of medical image segmentation were applied for evaluating the quality of experimental results. They are Dice score (DS), Jaccard index (JA), accuracy (AC), precision (PR), sensitivity (SE), specificity (SP), and average symmetric surface distance (ASSD [voxel]). Their definitions are given as follows:

$DS = \frac{2 \cdot N_{tp}}{2 \cdot N_{tp}+N_{fn}+N_{fp}}$, $JA = \frac{N_{tp}}{N_{tp}+N_{fn}+N_{fp}}$, $AC = \frac{N_{tp}+N_{tn}}{N_{tp}+N_{fn}+N_{fp}+N_{tn}}$, $PR = \frac{N_{tp}}{N_{tp}+N_{fp}}$, $SE = \frac{N_{tp}}{N_{tp}+N_{fn}}$,

$$SP = \frac{N_{tn}}{N_{tn}+N_{fp}}, \quad ASSD = \frac{\sum_{v_p \in S_p} d(v_p, S_g) + \sum_{v_g \in S_g} d(v_g, S_p)}{|S_p|+|S_g|}, \tag{19}$$

where $N_{tp}$, $N_{tn}$, $N_{fp}$ and $N_{fn}$ represent the number of pixels belonging to true positive, true negative, false positive and false negative, respectively. $S_p$ and $S_g$ denote the sets of surface voxels of predicted liver volume and ground truth, respectively. $d(v, S)$ denotes the minimum Euclidean distance between a voxel $v$ and surface $S$, and $|S|$ denotes the number of pixels belonging to the surface $S$. The format of mean ± standard deviation is used to present all the metrics for showing the average performance and the variations of results cross-subjects. In our experiments, the ranking of the performances was mainly based on the Dice score since it is the most important and the most commonly used one in medical image segmentation tasks.

**4.3. Implementation details**

The proposed unsupervised cross-modality domain adaptation framework was implemented in Python 3.6.5 with Pytorch 1.2.0 (Paszke et al., 2019). All deep learning experiments were performed on two NVIDIA TITAN V GPUs with 12 GB memory in parallel. The network U1 was pre-trained on the source CT dataset in a fully-supervised learning manner with Adam optimizer (Kingma and Ba, 2014) (the learning rate 0.001 initially and decreased by 10% every epoch; batch size 8). All the trainable networks U2, U3, U4, D1, and D2 in the proposed framework were trained with RMSprop optimizer (Graves, 2013) (smoothing constants 0.9; learning rates 0.0001, 0.00012, 0.00015, 0.0002, and 0.0002, respectively; batch sizes 8). The weights of the networks U1, U3, U4, D1, and D2 were initialized by Kaiming's normal method (He et al., 2015). The weights of discriminators D1 and D2 were clipped between -0.01 to 0.01 during training.

**4.4. Adversarial learning for domain adaptation of U2**

*4.4.1. Post-situ identification manner*

Note we defined two types of adversarial learning for domain adaptation named in-situ identification manner and post-situ identification manner according to the relative position of the discriminator and the features that need to align their distributions. Most previous studies leveraged the idea of the former one to achieve unsupervised domain adaptation. Note the role of U2 in our proposed framework is important for the final segmentation results since it provides the pseudo-label. To investigate the effectiveness of in-situ and post-situ identification manners on domain adaptation of U2, we designed two experiments: i) extract the part consisting of U1, U2, and D1 from the proposed framework as the post-situ identification manner (referred as PSIM); ii) abandon D1 and construct another discriminator D3 with similar structure to D1 to immediately follow the features whose distributions need to be aligned as the in-situ identification manner (referred as ISIM).

The quantitative comparison results are shown in Table 1. Note the performance of U2 under the setting ISIM is quite low with a Dice score of 0.237±0.116, while that of U2 under setting PSIM has much higher Dice score, Jaccard index, precision, sensitivity and average symmetric surface distance. The P-values of the paired t-test on the group of "PSIM vs ISIM" are smaller than 0.05 in all metrics in Table 2, showing significant differences. Fig. 13 shows several examples of segmentation results of U2 under the settings ISIM and PSIM. Note the segmentations under the setting ISIM completely missed the anatomical structure of the liver, while the segmentations under the setting PSIM indicated the location and outlined partial structure of the liver in most normal samples. These results demonstrated that the adversarial learning with post-situ identification manner achieves better domain adaptation.

Fig. 14 provides further insight of the in-situ and post-situ identification manners by looking into the training progress. Note the Dice score of validation results under the setting ISIM is always at a low level during the training period, which means the model U2 cannot learn much useful information. By contrast, the Dice score of validation results under the setting PSIM increases sharply with the increase of epoch during training, reaches the peak when epoch is about 30, and then slowly decreases. This implies that the model U2 can effectively learn task-related information in a relative correct direction.

Table 1 Quantitative comparison of segmentation performance of U2 between different settings of adversarial learning.

| Setting | DS | JA | AC | PR | SE | SP | ASSD | Training time |
|---|---|---|---|---|---|---|---|---|
| ISIM | 0.237±0.116 | 0.139±0.075 | 0.934±0.015 | 0.314±0.195 | 0.212±0.108 | 0.973±0.015 | 22.664±5.822 | 4341s |
| PSIM/SA | 0.759±0.182 | 0.634±0.158 | 0.977±0.016 | 0.753±0.189 | 0.774±0.194 | 0.989±0.004 | 3.315±3.140 | 2817s |
| SEA | 0.057±0.033 | 0.029±0.018 | 0.852±0.018 | 0.041±0.024 | 0.094±0.059 | 0.892±0.012 | 34.011±2.836 | 3589s |
| SA+SEA | **0.788±0.187** | **0.674±0.164** | **0.980±0.016** | **0.808±0.196** | **0.776±0.195** | **0.992±0.003** | **1.270±1.351** | 4502s |

Table 2 P-values of paired t-test on the segmentation performances of U2 between different settings of adversarial learning

| Group | DS | JA | AC | PR | SE | SP | ASSD |
|---|---|---|---|---|---|---|---|
| PSIM/SA vs ISIM | 0.000 | 0.000 | 0.000 | 0.000 | 0.000 | 0.000 | 0.000 |
| SA+SEA vs PSIM/SA | 0.001 | 0.001 | 0.000 | 0.000 | 0.731 | 0.000 | 0.003 |

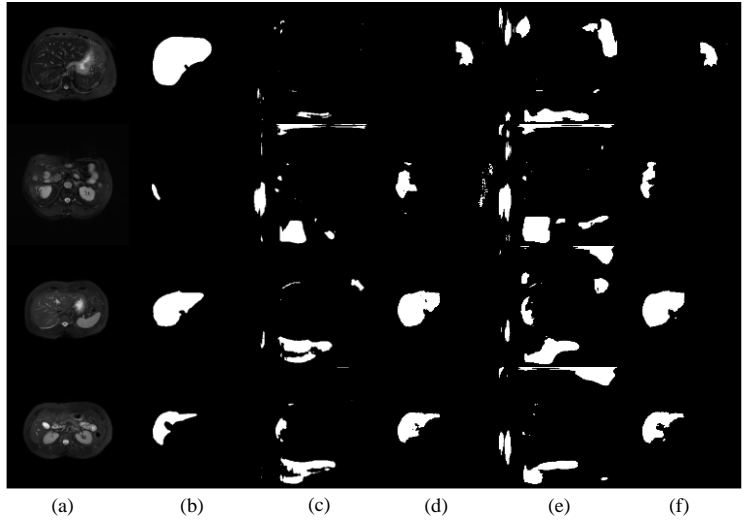

(a)　　　(b)　　　(c)　　　(d)　　　(e)　　　(f)

Fig. 13. Examples of segmentation results of U2 under different settings of adversarial learning. (a) MR image; (b) Ground truth; (c) Segmentation predicted under setting ISIM; (d) Segmentation predicted under setting PSIM/SA; (e) Segmentation predicted under setting SEA; (f) Segmentation predicted under setting SA+SEA. The first row is hard sample, and the next three rows are normal samples.

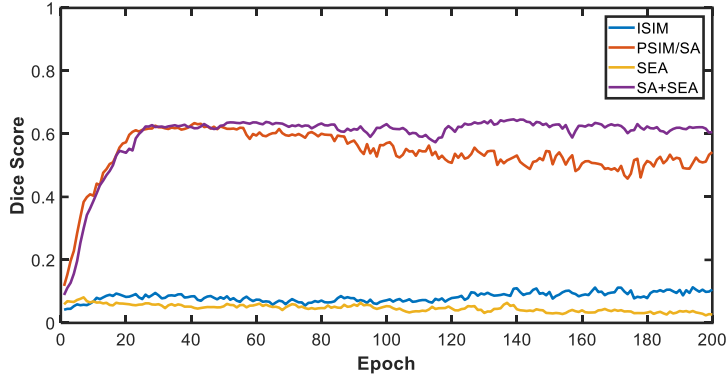

Fig. 14. (Best viewed in color) Per-epoch validation results under different settings of adversarial learning.

*4.4.2. Joint semantic-aware and shape-entropy-aware adversarial learning*

Previous studies either used semantic information (Chen et al., 2018) or entropy information (Vu et al., 2019) to achieve unsupervised domain adaptation. We integrate semantic information, entropy information, and shape information in joint semantic-aware and shape-entropy-aware adversarial learning to achieve unsupervised cross-modality domain adaptation. To verify the superiority of the proposed approach, three experiments were conducted, including: i) extracted the part of framework consisting of U1, U2, and D1 as the semantic-aware adversarial learning (referred as SA, same setting as PSIM); ii) extracted the part of framework consisting of U1, U2, and D2 as the shape-entropy-aware adversarial learning (referred as SEA); iii) extracted the part of framework consisting of U1, U2, D1, and D2 as the joint semantic-aware and shape-entropy-aware adversarial learning (referred as SA+SEA).

Table 1 shows the quantitative comparison results of SA, SEA, and SA+SEA. Note the network U2 achieved fairly good performance under the settings SA and SA+SEA with Dice scores of 0.759±0.182 and 0.788±0.187, respectively, while the segmentation performance of U2 under the setting SEA is quite low with a Dice score of 0.057±0.033. Note this is consistent with the results shown in Fig. 13. SEA cannot achieve the same high performance as the Vu et al. (2019) due to the more severe domain shift in our task. SA+SEA achieved the best performance, but slightly better than SA. The P-values of paired t-test on the group of "SA+SEA vs SA" are smaller than 0.05 in most metrics including Dice score in Table 2, showing significant differences. This implies the semantic information makes the most contributions to achieve domain adaptation, and the shape and entropy information only makes minor contribution. Integration of these three sets of information provides reasonable constraints for the model U2 and make the model learn in a correct direction. As Fig. 14 shows, compared with the setting SA, the validation results under the setting SA+SEA were more stable after reaching the peak, and did not show a downward trend during training.

## 4.5. Performance of the unsupervised domain adaptation framework

Note our proposed framework includes three networks U2, U3 and U4 that can be used to segment liver from MR image. In view of their interdependence, comparison of the performances of these three networks helps us better understand the whole framework and the desired model U3. The quantitative evaluations of the performances of U2, U3 and U4 are listed in Table 3. Overall, the network U2, U3 and U4 achieved fairly good performances on segmenting liver from MR image with Dice scores of 0.788±0.189, 0.912±0.037, and 0.897±0.084, respectively, for their respective roles in the proposed framework. Table 4 gives the paired t-test results of comparison of average performance of "U3 vs U2" and "U3 vs U4". Note the performance of U3 and U2 are significantly different (p-value < 0.05), while the performance of U3 and U4 are not significantly different (p-value > 0.05). Fig. 15 offers more detailed information about the segmentations of all subjects. The performance of U3 on each subject is better than that of U2, especially on the third subject whose slices are all hard samples. The performance of U3 on each subject is similar to that of U4, except for the third subject.

Fig. 16 shows the segmentation results. Note the network U2 cannot provide accurate and robust prediction of the structure of liver. U4 can accurately capture the liver structure in most normal samples and only can indicate the location of liver in most hard samples, while U3 can accurately capture the liver structure in most samples including most hard samples. These observations demonstrated the desired model U3 has the best segmentation accuracy and the most robust performance compared with U2 and U4.

Table 3 Quantitative evaluations on the performances of different networks in the proposed framework.

| Network | DS | JA | AC | PR | SE | SP | ASSD | Training time |
|---|---|---|---|---|---|---|---|---|
| U2 | 0.788±0.189 | 0.675±0.168 | 0.981±0.015 | 0.833±0.200 | 0.753±0.191 | 0.994±0.002 | 1.701±2.343 | 7770s |
| U3 (desired) | **0.912±0.037** | **0.840±0.060** | **0.991±0.005** | **0.931±0.024** | **0.897±0.070** | **0.997±0.001** | **0.214±0.239** | 7770s |
| U4 | 0.897±0.084 | 0.822±0.111 | 0.989±0.010 | 0.916±0.031 | 0.889±0.120 | 0.996±0.002 | 0.319±0.480 | 7770s |

Table 4 P-values of paired t-test on the performances of different networks in the proposed framework.

| Group | DS | JA | AC | PR | SE | SP | ASSD |
|---|---|---|---|---|---|---|---|
| U3 vs U2 | 0.003 | 0.000 | 0.001 | 0.040 | 0.000 | 0.000 | 0.011 |
| U3 vs U4 | 0.261 | 0.241 | 0.183 | 0.017 | 0.619 | 0.034 | 0.198 |

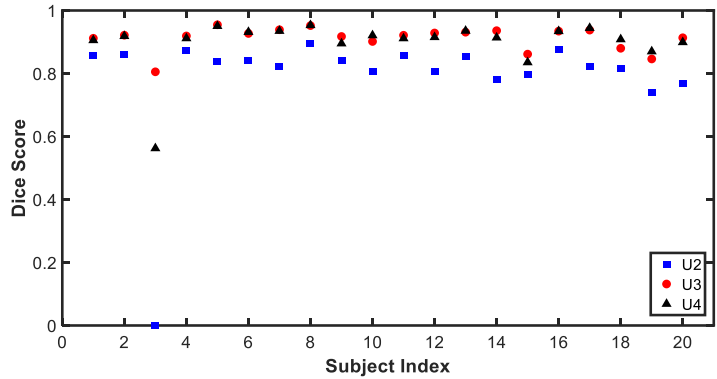

Fig. 15. Segmentation results of all subjects in target domain predicted by different networks in the proposed framework.

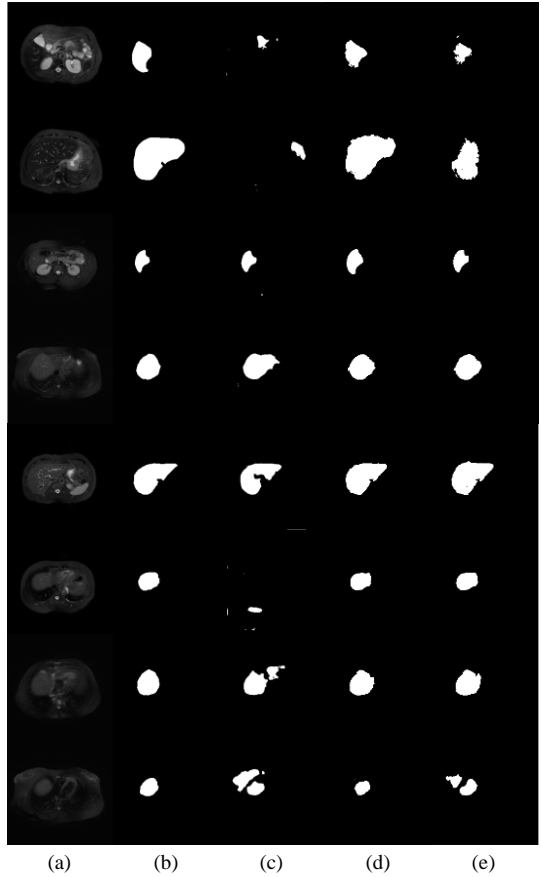

Fig. 16. Examples of segmentation results of different networks in the proposed framework. (a) MR image; (b) Ground truth; (c) Segmentation predicted by U2; (d) Segmentation predicted by the desired model U3; (e) Segmentation predicted by U4. The first two rows are hard samples from the third subject, and the last six rows are normal samples.

Fig. 17 illustrated what happened to these three networks during training. Note the Dice scores of U3 and U4 were lower than that of U2 before the 12$^{th}$ epoch, and then started to exceed that of U2. The Dice scores of the three networks show the same trend before the 70$^{th}$ epoch. They increased gradually to the peak value and then fluctuated around the peak. During this period, the Dice scores of U3 and U4 were comparable. At the 70$^{th}$ epoch, the Dice score of U3 increased sharply and was steadily higher than that of U4 after this point.

Note that the most instructive information at the beginning (before the 12$^{th}$ epoch) is the logit map and weighted self-information map generated by the pre-trained U1 with CT image in source domain. The proposed joint semantic-aware and shape-entropy-aware adversarial learning, which are based on the logit map and the weighted self-information map, are used to achieve domain adaptation of U2. In contrast, only semantic-aware adversarial learning, which is based on the logit map, can guide U3 to learn in a relative correct direction at the beginning when other information-based losses are not reliable yet (some of them may provide wrong learning directions). Therefore, it is reasonable that the Dice score of U2 is higher than U3 at the beginning.

As the performance of U2 continues to improve, the pseudo-label generated with the output of U2 for supervising U3 becomes more accurate, so the loss provided by it can guide U3 to learn in a more effective direction. With the improvement of the performance of U3, the predicted mask and the mask refined with pixel-adaptive mask refinement become more accurate, providing more instructive loss for supervising U3. Consequently, all four losses (one segmentation loss from U2, one adversarial loss from discriminator D1, one entropy minimization loss and one refinement loss from U3) form reasonable constraints to guide the desired model U3 to learn. This is why the performance of U3 started to surpass that of U2 after the 12$^{th}$ epoch.

U3 and U4 form a student-to-partner learning mode and the knowledge is transferred from the teacher model U3 to the student model U4 at the first stage (before the 70$^{th}$ epoch). The two models then learn from each other at the second stage (after the 70$^{th}$ epoch). Intuitively, the performance of U4 is expected to be lower than that of U3 all the time because the learning of U3 is guided by multiple constraints, while there is only one constraint for U4. However, the performance of U4 is almost the same as that of U3 before the 70$^{th}$ epoch, as Fig. 17 shows. The main reason is that the pseudo-label provided by U2 for U3 is not as accurate as the pseudo-label provided by U3 for U4. After the 70$^{th}$ epoch, the pseudo-label provided by U2 for U3 is replaced by the more accurate pseudo-label provided by U4, so the performance of U3 begins to exceed that of U4.

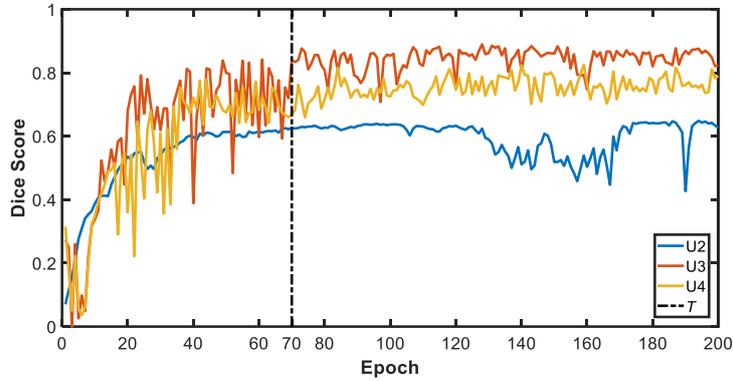

Fig. 17. (Best viewed in color) Per-epoch validation results of different networks in the proposed framework. $T$ denotes the time (epoch) that the student-to-partner learning module in the proposed framework enters the second stage.

**4.6. Ablation Study**

Recall from section 3.2 that there are six crucial components in our unsupervised cross-modality domain adaptation framework. The advantages of post-situ identification manner for adversarial learning and the joint semantic-aware and shape-entropy-aware adversarial learning have been verified in section 4.4. The effects of the rest components on the proposed framework are investigated as below.

*4.6.1. Mean completer of pseudo-label generation*

As mentioned in section 3.6, the pseudo-label $y_{2,t}$ in the Fig. 2 generated via the normal manner can fail when the input of U2 is hard sample. We proposed the mean completer of pseudo-label generation method to address this problem. Table 5 and Table 6 show the results of the performance if we remove it from the proposed framework (referred as "w/o MCPLG"). Note the performance remains basically unchanged on the whole without this component. To further understand the function of this component, we plot the segmentation results of all subjects with and without it in Fig. 18. Note the segmentation results are comparable between them except for the third subject whose slices are all hard samples. The Dice score increased from 0.761 to 0.805 after adding the mean completer. This evidences that the proposed mean completer of pseudo-label generation approach can indeed address the performance degradation caused by hard sample to a certain extent.

Table 5 Quantitative evaluations on the performances of the desired model U3 under different settings in ablation study.

| Setting | DS | JA | AC | PR | SE | SP | ASSD | Training time |
|---|---|---|---|---|---|---|---|---|
| Proposed | **0.912±0.037** | **0.840±0.060** | **0.991±0.005** | 0.931±0.024 | 0.897±0.070 | **0.997±0.001** | **0.214±0.239** | 7770s |
| w/o MCPLG | 0.909±0.045 | 0.836±0.071 | 0.991±0.006 | **0.933±0.031** | 0.889±0.075 | 0.997±0.002 | 0.226±0.223 | 7583s |
| w/o LSAF | 0.843±0.132 | 0.745±0.152 | 0.983±0.016 | 0.929±0.032 | 0.878±0.155 | 0.990±0.009 | 0.736±1.503 | 7534s |
| with PP | 0.904±0.036 | 0.827±0.058 | 0.990±0.005 | 0.914±0.026 | **0.898±0.070** | 0.996±0.001 | 0.218±0.215 | 7952s |
| w/o PAMR | 0.892±0.034 | 0.807±0.054 | 0.989±0.005 | 0.898±0.032 | 0.889±0.058 | 0.995±0.002 | 0.303±0.242 | 6914s |
| w/o STPL | 0.899±0.048 | 0.821±0.074 | 0.990±0.006 | 0.929±0.032 | 0.874±0.073 | 0.996±0.002 | 0.241±0.178 | 5843s |
| with DML | 0.893±0.047 | 0.809±0.073 | 0.989±0.007 | 0.918±0.038 | 0.872±0.072 | 0.996±0.002 | 0.285±0.296 | 8356s |
| w/o SSL | 0.871±0.049 | 0.775±0.072 | 0.987±0.007 | 0.925±0.035 | 0.830±0.086 | 0.996±0.002 | 0.309±0.176 | 5986s |

Table 6 P-values of paired t-test on the performances of the desired model U3 under different settings in ablation study

| Group | DS | JA | AC | PR | SE | SP | ASSD |
|---|---|---|---|---|---|---|---|
| Proposed vs w/o MCPLG | 0.357 | 0.397 | 0.388 | 0.781 | 0.167 | 0.812 | 0.813 |
| Proposed vs w/o LSAF | 0.014 | 0.004 | 0.010 | 0.002 | 0.466 | 0.003 | 0.130 |
| Proposed vs with PP | 0.000 | 0.000 | 0.000 | 0.000 | 0.195 | 0.000 | 0.557 |
| Proposed vs w/o PAMR | 0.001 | 0.000 | 0.000 | 0.000 | 0.295 | 0.000 | 0.003 |
| Proposed vs w/o STPL | 0.011 | 0.008 | 0.025 | 0.743 | 0.009 | 0.870 | 0.620 |
| Proposed vs with DML | 0.000 | 0.000 | 0.000 | 0.075 | 0.001 | 0.090 | 0.001 |
| Proposed vs w/o SSL | 0.000 | 0.000 | 0.000 | 0.487 | 0.000 | 0.735 | 0.054 |

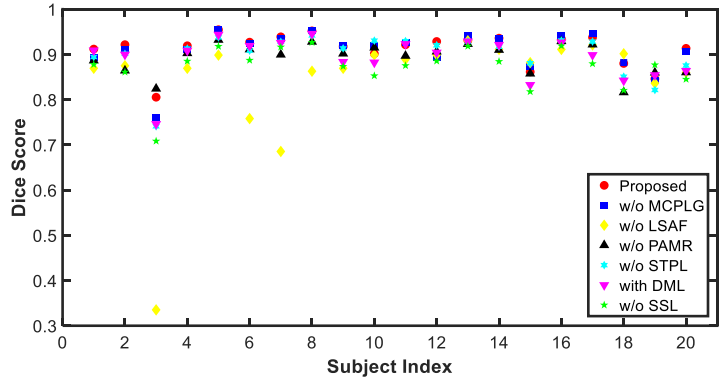

Fig. 18. (Best viewed in color) Segmentation results of all subjects in target domain under different settings in ablation study.

*4.6.2. Low-signal augmentation function*

We proposed the low-signal augmentation function to further improve the robustness of the desired model U3 in the presence of hard samples. Table 5 and Table 6 show the results of the performances with (denoted as "Propose") and without (denoted as "w/o LSAF") this process. Note the significant improvement in most metrics including Dice score by including this process.

In Fig. 18, the segmentation accuracy of most subjects under the setting "Proposed" is higher that under the setting "w/o LSAF", especially for the third, sixth, seventh, and eighth subjects. Their Dice scores drop markedly from 0.805, 0.927, 0.934, and 0.952 to 0.335, 0.758, 0.685, and 0.863, respectively, without the low-signal augmentation function. The third subject is the only one whose slices are all hard samples, which is the reason why its Dice score decreased the most. This verifies the effectiveness of the proposed low-signal augmentation function on enhancing the robustness of U3 in the presence of hard samples.

*4.6.3. Pixel-adaptive mask refinement*

Recall from section 3.8 that we use pixel-adaptive mask refinement as a self-learning method to further improve the performance of the desired model U3. Here we investigate two relevant issues: i) can pixel-adaptive mask refinement be used as a post-processing method to refine the predicted liver mask? ii) can this method improve the performance of U3 as a self-learning way?

As a local mask refinement method, similar to CRFs (Krähenbühl and Koltun, 2012), pixel-adaptive mask refinement is naturally used as a post-processing approach for semantic segmentation. Table 5 and Table 6 are the results of the performances with (denoted as "with PP") and without (denoted as "Proposed") such post-processing. Note the segmentation accuracy significantly decreases after including this post-processing. Fig. 19 can be used to assist understanding of this unexpected result. Note the boundaries of the refined segmentation became blurry after post-processing. We speculate that the main reason that the refined segmentations in our experiments are not as good as that in Araslanov and Roth (2020) is that the boundaries between organs in MR images are not as clear as those in natural images. Hence it is not recommended employing the pixel-adaptive mask refinement as the post-processing to refine the liver segmentations in our task.

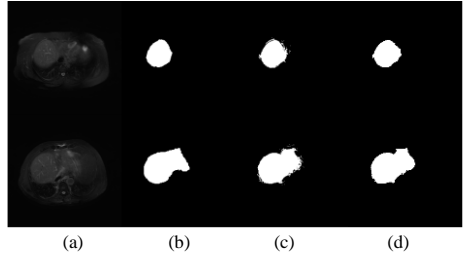

Fig. 19. Examples of the segmentations refined with pixel-adaptive mask refinement. (a) MR image; (b) Ground truth; (c) Refined segmentation; (d) Original segmentation.

To answer the second question, Table 5 and Table 6 are the results of the performances with (denoted as "Proposed") and without (denoted as "w/o PAMR") this approach. Note the significant improvement of the performance after including the proposed approach.

It seems counterintuitive that pixel-adaptive mask refinement decreases the performance of segmentation when used as a post-processing process, but can provide effective guidance for U3. To understand this, we compared the refinement loss $\mathcal{L}_{pamr}$ (Eq. 15) and the segmentation loss $\mathcal{L}_{gt}$ (not involved in training) used in fully-supervised learning provided by ground truth. The results were shown in Fig. 20. Note that although $\mathcal{L}_{pamr}$ and $\mathcal{L}_{gt}$ did not show strong a correlation in the whole training process ($R$ is 0.202), after certain iterations, the two began to show a strong correlation. For example, during the period of 2000 to 3000 iterations, the correlation coefficient is as high as 0.670. Strong correlation means that $\mathcal{L}_{pamr}$ and $\mathcal{L}_{gt}$ have similar trends, that is, the learning direction provided by $\mathcal{L}_{pamr}$ for the model is close to that provided by $\mathcal{L}_{gt}$. This is why the refinement loss is useful for guiding the learning of U3 and improving the performance.

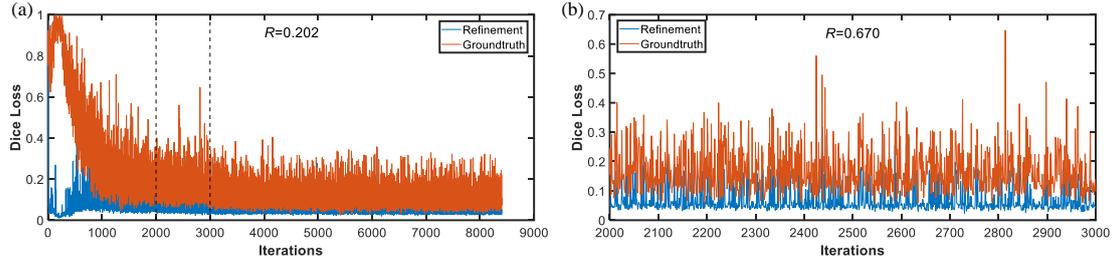

Fig. 20. (Best viewed in color) The refinement loss (involved in training) and the fully-supervised learning loss (not involved in training) of the desired model U3 during training. (a) The whole process of training; (b) Part of the training process. The bule line denotes the refinement loss, and the red line denotes the fully-supervised learning loss provided by ground truth. $R$ represents the correlation coefficient between the refinement loss and the fully-supervised learning loss.

*4.6.4. Student-to-partner learning*

The last crucial component in the proposed framework is student-to-partner learning. In this section, we investigate two issues: i) the effectiveness of student-to-partner learning to improve the performance of the desired model U3; ii) the superiority of student-to-partner learning over the deep mutual learning. For the former one, we removed the network U4 from the proposed framework and referred this setting as "w/o STPL". For the latter one, we designed a deep mutual learning module according to the structure in Fig. 12(a) by replacing the input of U4 with the transformed MR image $x_{t'}$ (see Fig. 2) and adding a segmentation loss provided by the pseudo-label $y_{2,t}$ (referred as "with DML"). The segmentation results and the corresponding paired t-test results under the above two settings are shown in Table 5 and Table 6.

Note the performance of U3 under the setting "w/o STPL" is significantly lower than that under the setting "Proposed" in terms of most metrics including Dice score. Fig. 18 also evidences this by showing that the Dice scores of most subjects under the setting "w/o STPL" are lower than that under the setting "Proposed". All the observations verify that the proposed student-to-partner learning can improve the performance of U3.

In terms of the deep mutual learning, we observed that the performance of U3 under the setting "with DML" is significantly lower than that under the setting "Proposed" in most metrics. The Dice scores of all subjects under the setting "with DML" are lower than that under the setting "Proposed" in Fig. 18. Those observations reveal that the deep mutual learning is not as good as the student-to-partner learning in improving the performance of U3. More interestingly, it is found that the performance of U3 under the setting "with DML" is also lower than that under the setting "w/o STPL" according to Table 5 Table 6, and Fig. 18. This shows that the deep mutual learning not only cannot improve the performance of U3, but decreases the performance to a certain extent. We speculate that the rough pseudo-label $y_{2,t}$ limits the performance of U4 in deep mutual learning module, and then limits the performance of U3 since these two networks learn from each other. The above observations verify that our student-to-partner learning is superior to deep mutual learning in our tasks when no ground truth is provided

The pixel-adaptive mask refinement and student-to-partner learning are employed as two self-learning methods to further improve the performance of the desired model U3. To investigate the effect of the combination of these two methods, we simultaneously removed the pixel-adaptive mask refinement and the network U4 from the proposed unsupervised domain adaptation framework (referred as "w/o SSL"). The segmentation results and the corresponding paired t-test results are shown in Table 5 and Table 6. Note that the performance of U3 under the setting "w/o SSL" is lower than that under the settings "Proposed", "w/o PAMR", and "w/o STPL". The Dice scores of all subjects also shows the same trend in Fig. 18. On the other hand, the performance of U3 without these two self-learning methods can still achieve a Dice score of 0.871±0.049. This reveals that the segmentation loss provided by the rough pseudo-label $y_{2,t}$, adversarial loss provided by the discriminator D1, and the entropy minimization loss already can guide the desired model U3 to learn in a fairly correct direction.

**4.7. Comparison with fully-supervised learning**

In this section, we compared the proposed unsupervised cross-modality domain adaptation method with fully-supervised learning methods (upper bound). To this end, four experiments were designed: i) trained and tested the model on the original MR images with labels in the target domain (referred as SL-O); ii) trained and tested the model on the MR images, which were transformed by the low-signal augmentation function, with labels in the target domain (referred as SL-T); iii) firstly trained the model on the labeled CT images in the source domain, and then fine-tuned and tested the model on the original MR images with labels (referred as STL-O); iv) firstly trained the model on the labeled CT images in the source domain, and then fine-tuned and tested the model on the transformed MR images with labels (referred as STL-T). The first two settings are supervised learning methods, and the last two are supervised transfer learning methods. In order to fairly compare the performances of the above four models and the desired model U3 in our unsupervised domain adaptation framework, these four models adopted the architecture of U3 and the hyperparameters of training U3. The results of the performances are shown in Table 7 and Table 8. Note the proposed method achieved comparable performance as four fully-supervised learning methods in terms of most metrics. In addition, the proposed method has obvious advantages in segmenting hard samples according to Fig. 21.

Table 7 Quantitative comparison of segmentation performance between our method and supervised learning-based methods

| Method | DS | JA | AC | PR | SE | SP | ASSD | Training time |
|---|---|---|---|---|---|---|---|---|
| Proposed | **0.912±0.037** | **0.840±0.060** | **0.991±0.005** | **0.931±0.024** | 0.897±0.070 | **0.997±0.001** | **0.214±0.239** | 7770s |
| SL-O | 0.902±0.152 | 0.843±0.167 | 0.990±0.015 | 0.922±0.083 | 0.896±0.176 | 0.996±0.002 | 0.272±0.651 | 820s |
| SL-T | 0.910±0.046 | 0.837±0.072 | 0.991±0.007 | 0.921±0.047 | **0.901±0.064** | 0.996±0.003 | 0.270±0.262 | 889s |
| STL-O | 0.894±0.172 | 0.834±0.179 | 0.990±0.016 | 0.914±0.111 | 0.886±0.188 | 0.996±0.002 | 0.534±1.632 | 826s |
| STL-T | 0.893±0.063 | 0.811±0.096 | 0.989±0.007 | 0.899±0.060 | 0.893±0.095 | 0.994±0.004 | 0.827±1.840 | 894s |

Table 8 P-values of paired t-test on the comparison of our method with supervised learning-based methods

| Group | DS | JA | AC | PR | SE | SP | ASSD |
|---|---|---|---|---|---|---|---|
| Proposed vs SL-O | 0.723 | 0.918 | 0.778 | 0.631 | 0.978 | 0.989 | 0.689 |
| Proposed vs SL-T | 0.592 | 0.652 | 0.394 | 0.342 | 0.575 | 0.253 | 0.374 |
| Proposed vs STL-O | 0.584 | 0.850 | 0.613 | 0.485 | 0.745 | 0.834 | 0.387 |
| Proposed vs STL-T | 0.086 | 0.098 | 0.097 | 0.011 | 0.826 | 0.009 | 0.153 |

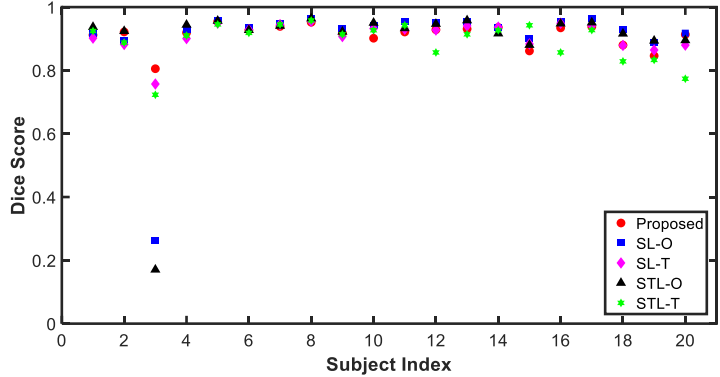

Fig. 21. (Best viewed in color) Segmentation results of all subjects in target domain with different methods.

## 5. Discussion

In the field of medical image analysis, domain adaptation is an important topic because of the diversity of medical data sources. The unsupervised domain adaptation is highly desirable as it is costly to obtain accurate medical image annotations. In previous methods, either adversarial learning or self-ensembling are used to address this problem. Our proposed unsupervised domain adaptation framework organically combines adversarial learning and self-learning methods and achieves comparable segmentation performance as four supervised learning-based methods (see Table 7 and Table 8). We conjecture that, in practice, there is a natural domain shift between the training and test datasets even they are both MR datasets. A simple supervised learning with only one single loss (even it is provided by ground truth) may suffer from this problem and falls into overfitting, making the segmentation accuracy on the test dataset lower than that on the training dataset. Multiple constraints (losses and/or regularizations) are usually employed to make the model to learn domain invariant representations and improve its generalization ability, such as multi-task learning (Ruder, 2017). We use multiple losses in the proposed unsupervised domain adaptation framework to guide the desired model U3 to learn more domain invariant representations during training, making the segmentation accuracy on the test dataset similar to that on the training dataset. Although the learning direction formed by those multiple losses is not as accurate as the direction in the aforementioned four supervised learning methods on the training dataset, the performance of the proposed unsupervised domain adaptation framework is comparable to these supervised learning methods on the test dataset.

In this paper, we defined two types of adversarial learning for unsupervised domain adaptation, in-situ and post-situ identification manners. The generator model and the discriminator model in in-situ identification manner are more difficult to learn than that in post-situ identification manner since the distributions contain more features in the former case. However, the features in shallow layers often contain a lot of tasks independent or even noise information. In in-situ identification manner, the generator will align the distributions of all these features indistinguishably. On the contrary, the post-situ identification manner can effectively guide the generator model to align the distributions of task-related features. This why the post-situ identification manner performed much better than the in-situ identification manner in our task (see Table 1, Table 2 and Fig. 13).

Most previous adversarial learning-based methods for unsupervised domain adaptation use the in-situ identification manner, which generally needs to align the distributions of middle/high-level features that contain more task-specific and semantic information compared with low-level features. They usually employ network without skip connections (such as residual networks or fully convolutional networks) to ensure the output of the back part of the network only relates to the features whose distributions need to be aligned (Chen et al., 2017; Dou et al., 2018; Hoffman et al., 2016). In the U-shaped network, it is not enough to just align the distributions of the features in the middle layers because the features in the shallow layers will also affect the output of the back part of the network through skip connections. If we select to align the distributions of the high-level features in deep layers, this will require updating almost all the weights of the network, which requires a large number of samples in the target domain. In addition, if the entire network is unfrozen, the U-shaped network as the generator may only focus on updating the weights of the decoding part for fooling the discriminator, which may lead to mode collapse, although we can use Wasserstein GAN to suppress it to some extent (Arjovsky et al., 2017). The proposed post-situ identification manner makes it possible to obtain domain adaptation effectively by aligning the feature distributions only in shallow layers, making the network focus on extracting low-level features. We extend the networks used in adversarial learning-based method to U-shaped networks which are most commonly used in medical image analysis. Aligning the feature distributions by only updating a small number of weights in shallow layers is conducive to avoiding mode collapse and coping with the common problems of small sample size in the target domain.

To the best of our knowledge, no previous published method has discussed the handling of hard samples when performing liver segmentation on MR images. The existence of hard samples can significantly deteriorate the segmentation performance. It can be seen from Fig. 21 that if the input are not the transformed MR images, the segmentation results of the subjects containing hard samples was quite poor even in a fully-supervised learning manner. We demonstrated the two techniques (mean completer of pseudo-label generation and low-signal augmentation function) can improve the segmentation accuracy of the desired model U3 in the presence of hard samples.

Deep mutual learning has achieved better performances than the traditional knowledge distillation method in a fully-supervised learning manner in several benchmark datasets (Zhang et al., 2018). However, in the field of unsupervised domain adaptation, only coarse cue (pseudo-label) rather than precise label (ground truth) is provided. If the deep mutual learning model is directly applied to unsupervised domain adaptation, the traditional "supervision learning" loss will be provided by the rough pseudo-label. This loss can effectively guide the two student models to learn useful knowledge theoretically in the initial stage. However, with the gradual improvement of segmentation, the pseudo-label cannot provide effective guidance for the models any more or even make the models learn in a wrong direction. In addition, the learning processes of the two student models are so similar that the outputs of the two models are very similar and the mimicry loss will be small, which may lead to the failure of mutual learning. We developed a novel learning mechanism named student-to-partner learning which overcomes these shortcomings and combines the advantages of traditional knowledge distillation and deep mutual learning. Its effectiveness on improving the performance of the desired model U3 in our unsupervised domain adaptation framework was demonstrated in section 4.6.4. We believe that the proposed student-to-partner learning can be effectively extended to other unsupervised domain adaptation tasks or even weakly supervised learning tasks.

## 6. Conclusion

In this work, we investigated a cross-modality liver segmentation problem from source labeled CT dataset to target unlabeled MRI dataset for reducing the cost of annotating medical images. We developed a novel unsupervised domain adaptation framework based on joint adversarial learning and self-learning. Our proposed framework comprises six crucial components. We validated the necessity and the effectiveness of these six components with extensive experiments. Using public data sets, we demonstrated our approach is robust on hard samples. The results show the proposed unsupervised domain adaptation approach achieved comparable performance as four supervised learning methods with a Dice score of $0.912 \pm 0.037$.

**Declaration of competing interest**

The authors declare that there are no conflicts of interest regarding the publication of this paper.

**CRediT authorship contribution statement**

**Jin Hong**: Conceptualization; Data curation; Formal analysis; Investigation; Methodology; Software; Validation; Visualization; Writing-original draft. **Simon Chun-Ho Yu**: Resources; Supervision; Writing-review & editing. **Weitian Chen**: Funding acquisition; Project administration; Resources; Supervision; Writing-review & editing.

**Acknowledgements**

This study was supported by a Faculty Innovation Award from the Faculty of Medicine of The Chinese University of Hong Kong, and a grant from the Innovation and Technology Commission of the Hong Kong SAR (Project MRP/046/20X).

We would like to acknowledge Dr. Liping Zhang and Mr. Yongcheng Yao for their valuable comments of this work.